\documentclass[runningheads]{llncs}

\usepackage{eccv}



\usepackage{eccvabbrv}

\usepackage{graphicx}
\usepackage{booktabs}
\usepackage{tikz}
\usepackage{stackengine}
\usepackage{color}

\usepackage[accsupp]{axessibility}  % Improves PDF readability for those with disabilities.

\usepackage[pagebackref,breaklinks,colorlinks,citecolor=eccvblue]{hyperref}

\usepackage{wrapfig}

\usepackage{orcidlink}

\usepackage{caption}
\begin{document}

% ---------------------------------------------------------------

\title{AI-Experiment: an augmented reality art project co-created with artificial intelligence}

\titlerunning{AI-Experiment}

\author{ Bernhard Egger\inst{1}\orcidID{0000-0002-4736-2397},
Hans Furer}%\inst{2}}

\authorrunning{Egger et al.}

 \institute{Friedrich-Alexander-Universität Erlangen-Nürnberg, Germany 
    \email{bernhard.egger@fau.de}\\%\and
   % FURER \& PARTNER Rechtsanwälte, Switzerland\\
    \email{hans.furer@furer-partner.ch}
    }

\maketitle

\begin{abstract}
We describe the creation of a physical artwork that was performed in synergy between an artist, a computer scientist and generative AI tools. A latent diffusion model was finetuned on photographs of the artwork of the specific artist and the generative model then generated synthetic artworks in a similar style. From the generated artworks a subset was selected to be painted again with acrylic on canvas by the original artist. In addition to the physical artifact an augmented reality component was added which brings the artwork to life with a set of generative video models. Together with this paper we release part of the artwork (training data), the code that was deployed to generate the artwork as well as the augmented reality experience for arts, research and education under \url{https://github.com/BernhardEgger/ARTimate}.
\end{abstract}

\section{Introduction}
\label{sec:intro}
\begin{wrapfigure}[20]{r}{0.4\textwidth}
    \vspace{-0.88cm}
        \includegraphics[width=\linewidth, keepaspectratio]{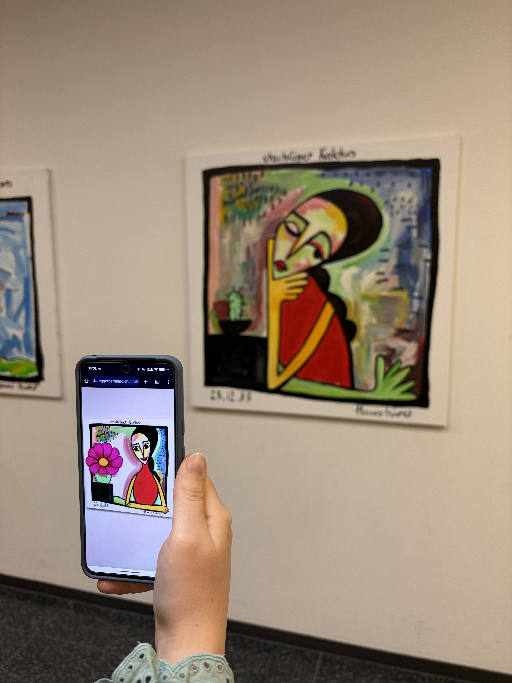}% 
  \caption{The augmented reality webapp recognizes the painting and plays the corresponding video.}
  \label{fig:app}
\end{wrapfigure}
Learning-based generative image models do have a long history. While simple PCA-based approaches did enable the generation of novel faces \cite{eigenfaces:sirovich1987low}, the first popular generative models that enabled more general and photorealistic image synthesis were generative adversarial networks \cite{gan:DBLP:journals/corr/GoodfellowPMXWOCB14}. These models also sparked interest among artists. To the best of our knowledge these models could not successfully generate more abstract artistic work, although they performed very well when trained on frontal faces \cite{stylegan2:DBLP:conf/cvpr/KarrasLAHLA20} and this attracted interest both within and outside the research community. Today's learning-based image generation models are heavily based on latent diffusion models \cite{stablediffusion:DBLP:conf/cvpr/RombachBLEO22}. They already in the very beginning started to outperform generative adversarial networks, not only on photorealism, but also on generation of diverse styles including more abstract styles. This raised the research question of whether they could generate abstract artwork that would be interesting to humans.
Whilst the research community and artists started to generate artwork in various styles that must have been represented in the training data, this procedure did not allow studying the research question in a controlled way. The release of DreamBooth \cite{dreambooth:DBLP:conf/cvpr/RuizLJPRA23}, the idea to finetune such models on a small novel dataset then finally triggered our idea to adapt such a generative model for our purpose of studying the capability to \textit{understand} abstract concepts. For the first ideas around this question we thought about experimenting with ArtBench \cite{DBLP:journals/corr/abs-2206-11404} which is a dataset of 10 distinct artistic styles with 6'000 images per style. We chose not to use the dataset since the copyright perspective of the dataset was unclear to us. We then decided to work on a smaller dataset of roughly 500 images but by one specific artist named Hans Furer who is a co-author of this paper, which was favorable for a variety of reasons:
\begin{itemize}
    \item He is alive and interested in new technologies including artificial intelligence
    \item Collaborating with the artist would eliminate copyright concerns
    \item The artist can be actively involved in the project and in this particular project also bring the generated artwork back to canvas
    \item The artwork of Hans Furer is very continuous, with a piece almost every month over 50 years and since 40 years in a square format which is ideal for finetuning a generative model
    \item The style of the artist is over time diverse enough that different aspects can be studied
\end{itemize}
In this paper we describe the process how five pairs of physical artwork on canvas were co-created.
The artwork still poses interesting questions in terms of intellectual property as well as the copyright of the artwork generated through this project. Whilst in this article we focus more on the technical aspects we are currently releasing a book \cite{egger2026kann} which puts the work in the broader context of digital arts and covers the legal aspects of the resulting work.

Following the same initial research question whether generative AI is capable to generate abstract art, we also investigated this in a different direction. From the moment the original artwork was finished on canvas and on exhibit at the university, we intended to bring it to life. This idea was inspired by \textit{The Borderline} exhibit at MIT. In the tunnels of MIT there is a 61 meter long wall with murals on it. The part that inspired us was that those murals are not only painted on the walls, but by downloading a mobile phone app, they can be brought to life and were animated in very creative ways. The artwork as well as the animations were created in 2017 by MIT affiliated artists, most likely without the use of generative AI. Our goal therefore was to bring our artworks to life in a similar style with state-of-the-art augmented reality tools and generative video models. 

From our perspective, the original AI-Experiment is unique since a generative AI was trained or finetuned to generate artwork in the style of one specific artist in combination with bringing the work actually back on canvas through the hands of the artist that created the training data. We did not find prior work that closes this loop.

\section{Related Work}
\label{sec:related}
From our perspective the most relevant work that is related to our project is \textit{The next Rembrandt}, where algorithms extracted various different features from 3D scans of paintings by Rembrandt to then generate a new piece which was then 3D printed\footnote{\url{https://www.vml.com/work/next-rembrandt}}. The main difference from our work lies in the synergy between the artist and the generative model. The work was co-created and inspired the artist which adds a unique component to our work. Another major difference lies in the deployed algorithms and also the underlying data. In our case, the algorithm was trained in photographs rather than 3D scans and the physical aspects of the artwork were not targeted during synthesis and this component was left to the artistic freedom of the artist.

The augmented reality component was strongly inspired by \textit{The Borderline} exhibit at MIT where the animations to the best of our knowledge were designed by artists rather than generated through generative models. The manipulation and animation of artwork with computer vision and computer graphics algorithms has a long history including works that build on 3D models \cite{3dmm:DBLP:conf/siggraph/BlanzV99}. In recent years generative video models have been used by artists to animate mostly classical artworks and also in research papers to demonstrate the capabilities of the model. Most of the work at the time we performed our experiments focused on near photorealistic art that is well represented in the training data of the models as well as mostly on faces.

For the augmented reality component various different apps including Artivive and Adobe Aero already existed at the time. Many of those apps however do not work on all mobile platforms in their most recent versions or were discontinued which motivated us to use a browser-based augmented reality toolkit instead. 

\section{Data}
\label{sec:data}
\begin{figure}[htbp]
    \centering
    \setlength{\lineskip}{1pt}
    \setlength{\parskip}{0pt}
% Calculate exact horizontal gap so 11 images span the line
\newlength{\imggap}%
\setlength{\imggap}{\dimexpr(\textwidth - 11\dimexpr0.080\textwidth\relax)/10\relax}%

% Explicit list of all 131 filenames in train/1981-1990/
\def\myfilenames{TRD_11_001-01_116,TRD_11_001-01_117,Furer_67-S000,TRD_11_001-01_118,TRD_11_001-01_119,TRD_11_001-01_102,TRD_11_001-01_113,TRD_11_001-01_114,TRD_11_001-01_115,TRD_11_001-01_107,TRD_11_001-01_108,TRD_11_001-01_109,TRD_11_001-01_110,TRD_11_001-01_112,TRD_11_001-01_103,TRD_11_001-01_104,TRD_11_001-01_105,TRD_11_001-01_106,TRD_11_001-01_018,TRD_11_001-01_135,Dame,TRD_11_001-01_136,TRD_11_001-01_137,TRD_11_001-01_138,TRD_11_001-01_120,Furer_68-S000,TRD_11_001-01_133,TRD_11_001-01_134,TRD_11_001-01_130,TRD_11_001-01_131,TRD_11_001-01_132,TRD_11_001-01_121,TRD_11_001-01_127,TRD_11_001-01_128,TRD_11_001-01_129,TRD_11_001-01_123,TRD_11_001-01_124,TRD_11_001-01_125,TRD_11_001-01_126,TRD_11_001-01_142,TRD_11_001-01_143,TRD_11_001-01_144,TRD_11_001-01_025,TRD_11_001-01_145,TRD_11_001-01_146,TRD_11_001-01_057,TRD_11_001-01_139,TRD_11_001-01_140,TRD_11_001-01_149,TRD_11_001-01_150,TRD_11_001-01_147,TRD_11_001-01_148,TRD_11_001-01_024,TRD_11_001-01_151,TRD_11_001-01_152,TRD_11_001-01_153,TRD_11_001-01_154,TRD_11_001-01_055,TRD_11_001-01_155,TRD_11_001-01_156,TRD_11_001-01_028,TRD_11_001-01_019,TRD_11_001-01_157,TRD_11_001-01_158,TRD_11_001-01_444,TRD_11_001-01_159,TRD_11_001-01_160,TRD_11_001-01_161,TRD_11_001-01_162,TRD_11_001-01_163,TRD_11_001-01_021,TRD_11_001-01_164,Furer_71-S000,TRD_11_001-01_165,TRD_11_001-01_167,TRD_11_001-01_166,TRD_11_001-01_027,Furer_72-S000,TRD_11_001-01_482,TRD_11_001-01_122,TRD_11_001-01_170,TRD_11_001-01_171,TRD_11_001-01_172,TRD_11_001-01_169,TRD_11_001-01_175,Furer_73-S000,TRD_11_001-01_173,TRD_11_001-01_168,TRD_11_001-01_458,TRD_11_001-01_184,TRD_11_001-01_183,TRD_11_001-01_182,TRD_11_001-01_188,TRD_11_001-01_187,TRD_11_001-01_186,TRD_11_001-01_191,TRD_11_001-01_202,TRD_11_001-01_190,TRD_11_001-01_189,TRD_11_001-01_180,TRD_11_001-01_181,TRD_11_001-01_185,TRD_11_001-01_197,TRD_11_001-01_196,TRD_11_001-01_195,TRD_11_001-01_198,Furer_75-S000,TRD_11_001-01_193,TRD_11_001-01_192,TRD_11_001-01_201,TRD_11_001-01_483,TRD_11_001-01_200,TRD_11_001-01_205,TRD_11_001-01_199,TRD_11_001-01_204,TRD_11_001-01_203,TRD_11_001-01_210,Furer_76-S000,TRD_11_001-01_016,TRD_11_001-01_209,TRD_11_001-01_207,TRD_11_001-01_463,TRD_11_001-01_206,TRD_11_001-01_017,TRD_11_001-01_213,TRD_11_001-01_212,TRD_11_001-01_211,TRD_11_001-01_216,TRD_11_001-01_215,TRD_11_001-01_214,TRD_11_001-01_460}

\begingroup
\raggedright%
\parfillskip=0pt plus 1fil%
\offinterlineskip%
\noindent%
\foreach \fname [count=\i] in \myfilenames {%
    \includegraphics[width=0.080\textwidth]{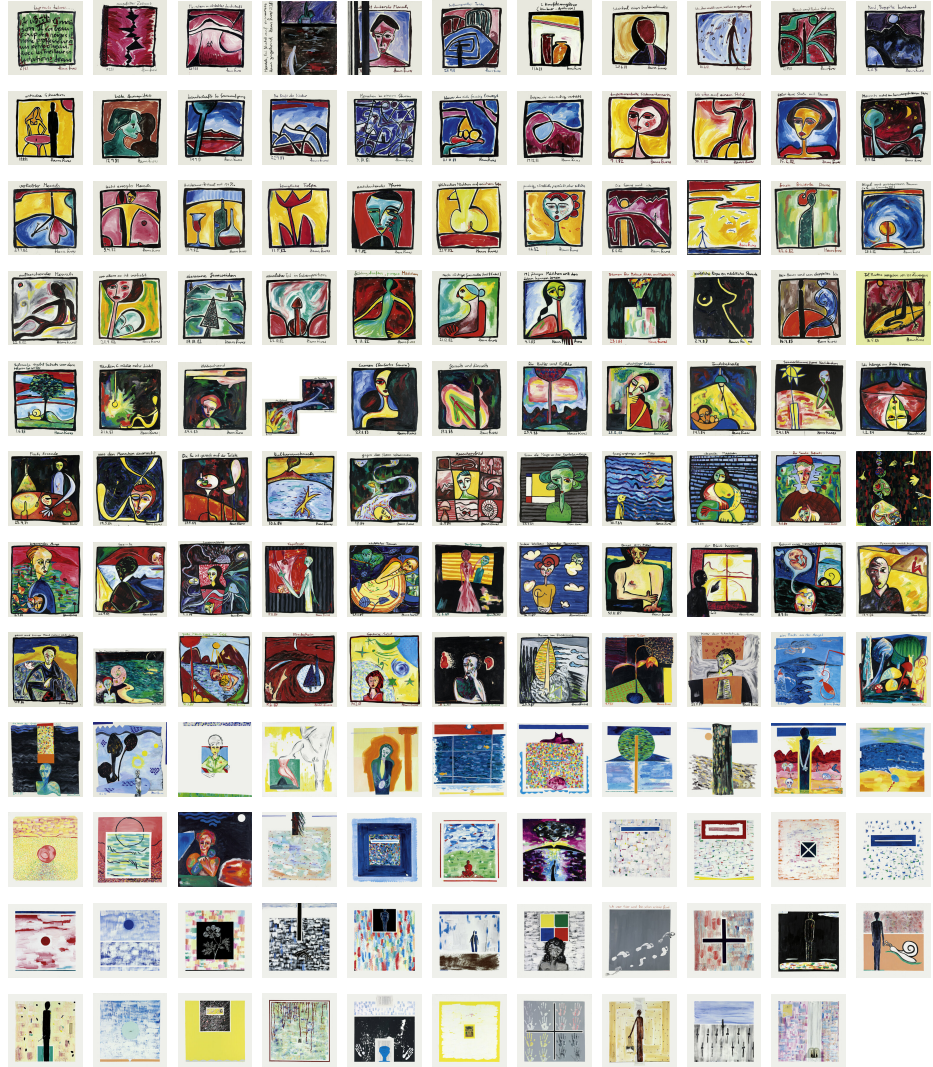}%
    \pgfmathtruncatemacro{\colmod}{mod(\i, 11)}%
    \ifnum\colmod=0 %
        \par\vspace{6pt}\noindent%
    \else%
        \hspace{\imggap}%
    \fi%
}%
\par
\endgroup
    \caption{131 paintings by Hans Furer that form the training data for the decade from 1981-1990. In this decade there are reoccurring topics and many drawings of human faces and bodies. }
    \label{fig:train80}
\end{figure}

\begin{figure}[htbp]
    \centering
    \setlength{\lineskip}{1pt}
    \setlength{\parskip}{0pt}

% Calculate exact horizontal gap so 11 images span the line
\setlength{\imggap}{\dimexpr(\textwidth - 11\dimexpr0.080\textwidth\relax)/10\relax}%
% Explicit list of all 131 filenames in train/1991-2000/
\def\myfilenames{Furer_77-S097,TRD_11_001-01_218,TRD_11_001-01_217,TRD_11_001-01_046,TRD_11_001-01_293,TRD_11_001-01_224,TRD_11_001-01_223,TRD_11_001-01_051,TRD_11_001-01_222,TRD_11_001-01_219,TRD_11_001-01_368,TRD_11_001-01_471,Furer_78-S099,TRD_11_001-01_227,TRD_11_001-01_228,TRD_11_001-01_020,TRD_11_001-01_226,TRD_11_001-01_225,TRD_11_001-01_229,TRD_11_001-01_221,TRD_11_001-01_472,TRD_11_001-01_220,TRD_11_001-01_239,TRD_11_001-01_238,TRD_11_001-01_237,TRD_11_001-01_236,Furer_79-S101,TRD_11_001-01_240,TRD_11_001-01_015,TRD_11_001-01_235,TRD_11_001-01_234,TRD_11_001-01_480,TRD_11_001-01_232,TRD_11_001-01_233,TRD_11_001-01_231,TRD_11_001-01_242,TRD_11_001-01_230,TRD_11_001-01_241,_HT23789,TRD_11_001-01_247,TRD_11_001-01_023,TRD_11_001-01_246,Furer_80-S101,TRD_11_001-01_245,TRD_11_001-01_244,TRD_11_001-01_249,TRD_11_001-01_248,TRD_11_001-01_467,TRD_11_001-01_243,TRD_11_001-01_022,TRD_11_001-01_254,TRD_11_001-01_253,TRD_11_001-01_252,TRD_11_001-01_255,TRD_11_001-01_054,TRD_11_001-01_251,TRD_11_001-01_250,TRD_11_001-01_258,Furer_81-S000,TRD_11_001-01_257,TRD_11_001-01_256,TRD_11_001-01_263,TRD_11_001-01_262,TRD_11_001-01_266,TRD_11_001-01_265,TRD_11_001-01_264,TRD_11_001-01_261,TRD_11_001-01_268,Furer_82-S101,TRD_11_001-01_267,TRD_11_001-01_260,TRD_11_001-01_259,TRD_11_001-01_470,TRD_11_001-01_274,TRD_11_001-01_273,TRD_11_001-01_278,TRD_11_001-01_277,Furer_83-S000,TRD_11_001-01_276,TRD_11_001-01_005,TRD_11_001-01_007,TRD_11_001-01_275,TRD_11_001-01_272,TRD_11_001-01_271,TRD_11_001-01_270,TRD_11_001-01_269,TRD_11_001-01_285,TRD_11_001-01_284,TRD_11_001-01_026,TRD_11_001-01_283,TRD_11_001-01_282,TRD_11_001-01_286,TRD_11_001-01_281,Furer_85-S000,TRD_11_001-01_280,TRD_11_001-01_461,TRD_11_001-01_288,TRD_11_001-01_279,TRD_11_001-01_287,TRD_11_001-01_466,TRD_11_001-01_045,TRD_11_001-01_043,TRD_11_001-01_292,TRD_11_001-01_067,TRD_11_001-01_065,TRD_11_001-01_066,Furer_87-S000,TRD_11_001-01_048,TRD_11_001-01_289,TRD_11_001-01_291,TRD_11_001-01_179,TRD_11_001-01_174,TRD_11_001-01_111,TRD_11_001-01_290,TRD_11_001-01_077,Furer_88-S000,TRD_11_001-01_294,TRD_11_001-01_297,TRD_11_001-01_296,TRD_11_001-01_295,TRD_11_001-01_300,TRD_11_001-01_299,TRD_11_001-01_298,TRD_11_001-01_303,TRD_11_001-01_011,TRD_11_001-01_302,TRD_11_001-01_306,TRD_11_001-01_301,TRD_11_001-01_305,TRD_11_001-01_304,TRD_11_001-01_468}%
\begingroup
\raggedright%
\parfillskip=0pt plus 1fil%
\offinterlineskip%
\noindent%
\foreach \fname [count=\i] in \myfilenames {%
    \includegraphics[width=0.080\textwidth]{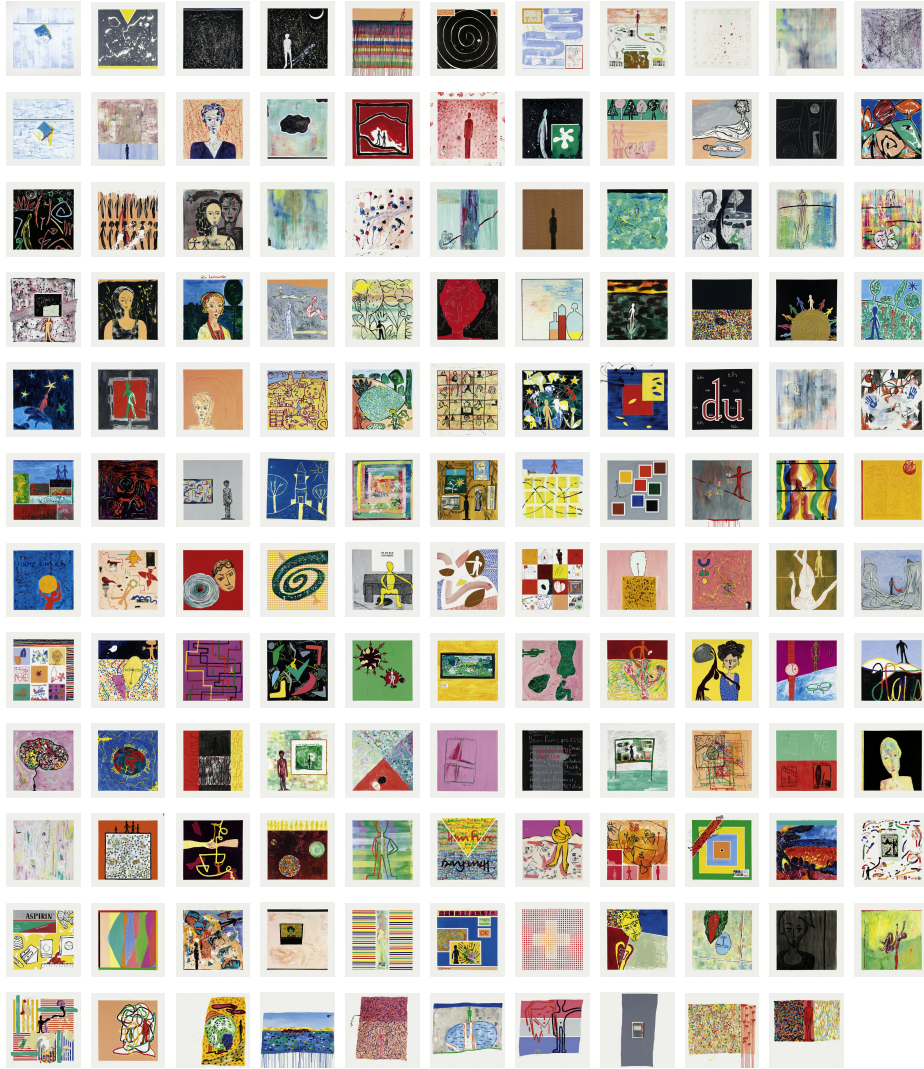}%
    \pgfmathtruncatemacro{\colmod}{mod(\i, 11)}%
    \ifnum\colmod=0 %
        \par\vspace{6pt}\noindent%
    \else%
        \hspace{\imggap}%
    \fi%
}%
\par
\endgroup

    \caption{131 paintings by Hans Furer that form the training data for the decade from 1991-2000. The artwork differs from the prior decade in color choice and also with a more abstract style.}
    \label{fig:train90}
\end{figure}

The artwork of Hans Furer was readily available in digital form since the artist published a catalogue raisonné in 2013 and Bernhard Egger designed an accompanying webpage\footnote{\url{https://hansfurer.com/}} to the catalogue. For the catalogue the photographer Tobias  R. Dürring captured high-quality digital photographs of the artwork. The artwork was divided into five decades (1971-1980, 1981-1990, 1991-2000, 2001-2010, 2011-2023) with each decade consisting of between 131 and 201 paintings (roughly one painting per month). Starting from 1980 the format of the paintings is consistently 100 x 100 cm executed in acrylic on canvas. Together with this publication we release a part of this dataset from 1981-1990 under a permissive license for artistic, research and educational use. Data from additional years can be shared upon reasonable request. The training data for two decades is shown in Figure~\ref{fig:train80} and Figure~\ref{fig:train90}.

%201 131 131 147 131

\section{Methods}
\label{sec:methods}

\subsection{Image Generation}
\label{sec:imagegen}

Image generation was performed based on DreamBooth \cite{dreambooth:DBLP:conf/cvpr/RuizLJPRA23}, a method to finetune a latent diffusion model to a particular object or identity. We based our implementation on the DreamBooth implementation by Shivam Shrirao\footnote{https://github.com/ShivamShrirao/diffusers} which is based on \cite{von-platen-etal-2022-diffusers} but in addition provides a straightforward Jupyter notebook that was deployed on Google Colab. As pre-trained models we choose Stable Diffusion 1.5 as well as Stable Diffusion 2.1 as both of them are favorable depending on the decade we are training on. This also matches the experience many artists and researchers had, that depending on the specific goal the specific choices of the two models (e.g. the encoder and training diet) lead to different behavior.
In our case we did not aim to reproduce a specific identity or object but rather a specific style of an artist but the DreamBooth fine-tuning strategy still performed very well. We changed two parameters of the implementation. First we removed the prior preservation. This strategy usually ensures that the model is still able to generalize after finetuning; however, in our case, it leads to the undesired effect that the generated images are not adopting the style of the artist (as the loss forces it to generate images in the original style). Due to this change our model however also misses generalization capabilities, which means that since we are only training a comparable small set of images without language annotation, the model also looses its capability to generate images that fit a specific text prompt. We view this as feature, as the image generation for our model was not biased by a human induced text prompt. The second small change we made to better adopt the style was to increase the maximal number of training steps from 800 to 10'000 which leads to better style adoption.
For both those changes we did not perform extensive qualitative or quantitative evaluation but rather tuned them based on our own qualitative observation on a small subset of generated images. The release of part of our training data enables a more structured comparison of different and newer models as well as more modern fine-tuning techniques. We did not perform such an evaluation as it was not the focus of this project, we do however see already challenges in the choice or design of metrics to evaluate generated artwork in terms of style and content.
Whilst for our photographs proper color calibration is available, Stable Diffusion does not distinguish color spaces and assumes sRGB. For our specific project this is not a limitation, as the generation through the artists anyway adds a creative component with some artistic freedom.
\subsection{Realization}
\label{sec:realization}

\begin{figure}[htbp]
  \centering
\includegraphics[width=0.8\textwidth, keepaspectratio]{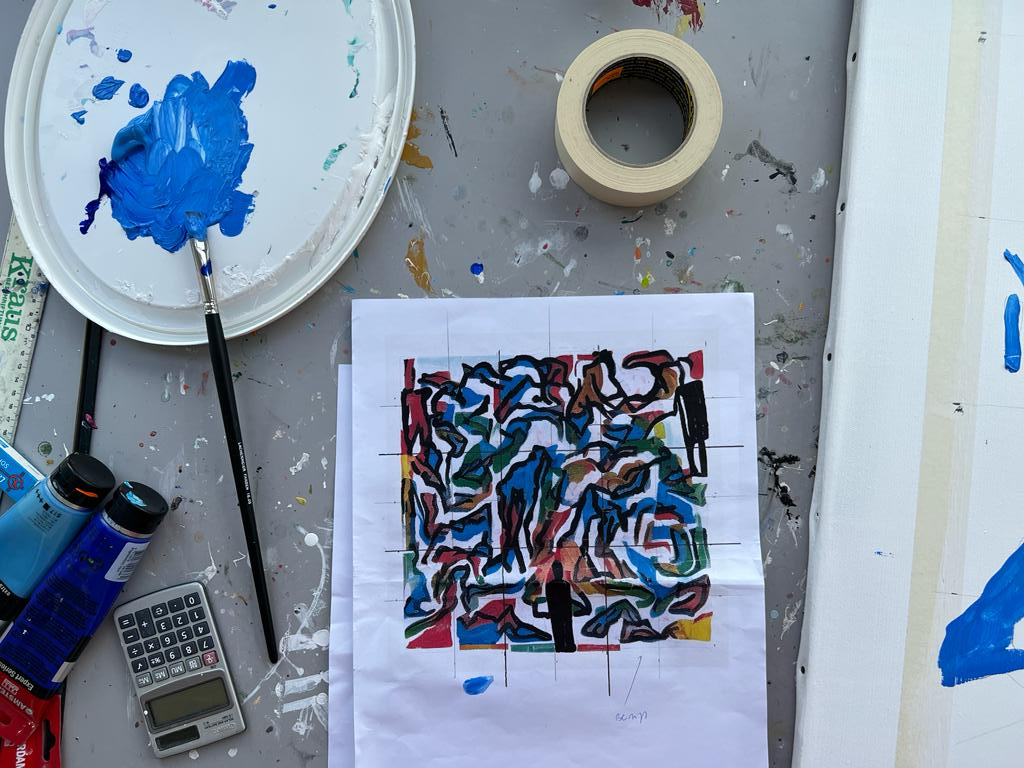}
  \caption{In the process of copying the generated image from a printed A4 sheet with acrylic to a 100 x 100 cm canvas.}
  \label{fig:copying}
\end{figure}

We generated 100 images for each of the variants of Stable Diffusion and for each decade. Bernhard Egger and Hans Furer then selected one representative generated piece that was painted by Hans Furer onto canvas. In addition, one sample from the training dataset was chosen in a way that it fits the generated piece very well (except for the third image, were the artist selected a particularly \textit{bad} piece from AI and a \textit{good} one of his own). The artist printed the image files with a standard printer on A4 paper and generated a replica of them step by step. He used a pencil to get the proportions as accurate as possible, see Figure~\ref{fig:copying} to get a gist of this process. The artist for this step, besides the motivation of a perfect copy, still had lots of freedom, which exact color to use and also how much of it to apply in which regions and with what structure.

\subsection{Video Generation}
\label{sec:videogen}
For video generation we relied on commercial and open weight models and either provided the painting as starting frame, or the generated painting as starting frame and the painting from the training data as end frame. In addition, the models were provided with text prompts. The models used were Kling~2.1, Veo~3, Wan~2.2, Hailuo~2.

\begin{figure}[htbp!]
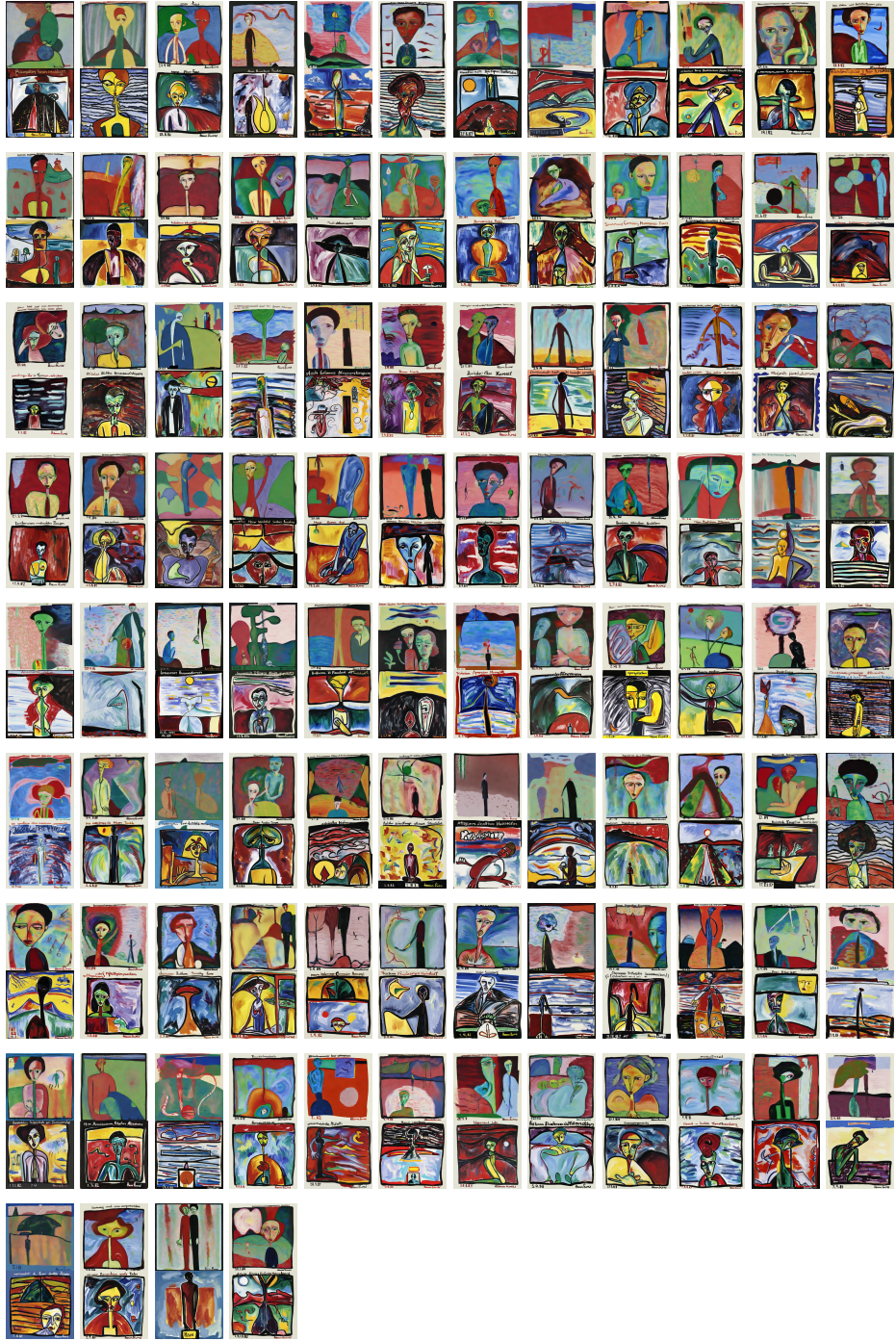

    \centering
    \setlength{\lineskip}{1pt}
    \setlength{\parskip}{0pt}
    \begingroup
    % Calculate exact horizontal gap so 12 images span the line
    \setlength{\imggap}{\dimexpr(\textwidth - 12\dimexpr0.076\textwidth\relax)/11\relax}%

\foreach \r in {0,...,8}{%
        % --- Index Calculations ---
        \pgfmathtruncatemacro{\startidx}{\r * 12}%
        \pgfmathtruncatemacro{\endidx}{min(\startidx + 11, 99)}%
        % --- Row from v1 ---
        \noindent\hbox to \textwidth{%
            \foreach \i in {\startidx,...,\endidx}{%
                \includegraphics[width=0.076\textwidth]{generated/v1/1981-1990/\i.png}%
                \ifnum\i=\endidx\else\hspace{\imggap}\fi%
            }%
        }\par\nointerlineskip%
        % --- Row from v2 ---
        \noindent\hbox to \textwidth{%
            \foreach \i in {\startidx,...,\endidx}{%
                \includegraphics[width=0.076\textwidth]{generated/v2/1981-1990/\i.png}%
                \ifnum\i=\endidx\else\hspace{\imggap}\fi%
            }%
        }\par\nointerlineskip%
        \ifnum\r<8 \vspace{6pt}\fi%
    }%
    \endgroup
     \vspace{6pt}
    \caption{We present the generated images by Stable Diffusion 1.5 (first rows) and Stable Diffusion 2.1 (second rows) for the decade from 1981-1990. Since the seed is kept constant it leads to very similar generations. We observe that for Stable Diffusion 2.1 there is much more detail and the style is better met.}
    \label{fig:generated_samples_v1_v280}
\end{figure}

\begin{figure}[htbp!]
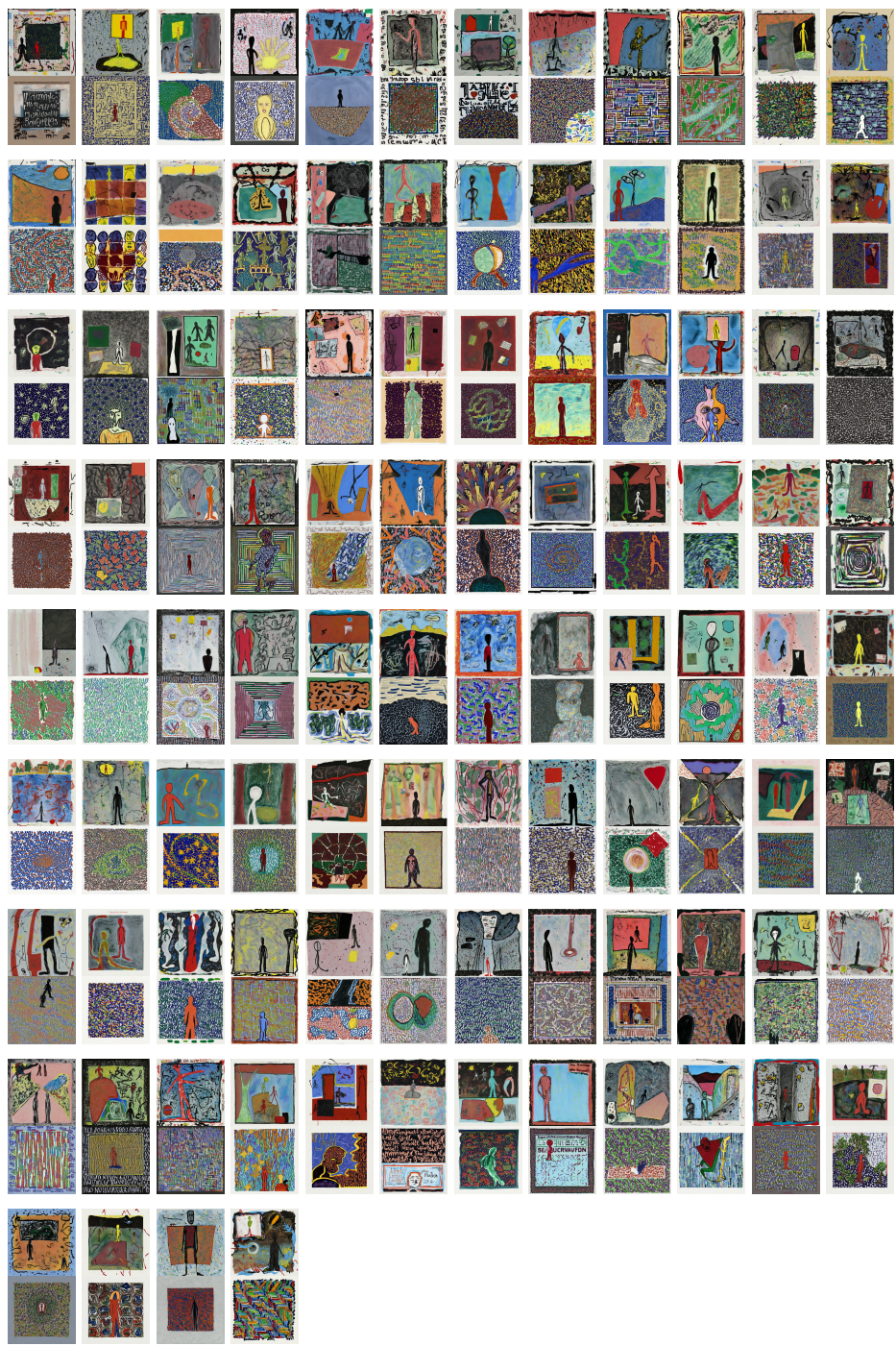

    \centering
    \setlength{\lineskip}{1pt}
    \setlength{\parskip}{0pt}
    \begingroup
    % Calculate exact horizontal gap so 12 images span the line
    \setlength{\imggap}{\dimexpr(\textwidth - 12\dimexpr0.076\textwidth\relax)/11\relax}%

\foreach \r in {0,...,8}{%
        % --- Index Calculations ---
        \pgfmathtruncatemacro{\startidx}{\r * 12}%
        \pgfmathtruncatemacro{\endidx}{min(\startidx + 11, 99)}%
        % --- Row from v1 ---
        \noindent\hbox to \textwidth{%
            \foreach \i in {\startidx,...,\endidx}{%
                \includegraphics[width=0.076\textwidth]{generated/v1/1991-2000/\i.png}%
                \ifnum\i=\endidx\else\hspace{\imggap}\fi%
            }%
        }\par\nointerlineskip%
        % --- Row from v2 ---
        \noindent\hbox to \textwidth{%
            \foreach \i in {\startidx,...,\endidx}{%
                \includegraphics[width=0.076\textwidth]{generated/v2/1991-2000/\i.png}%
                \ifnum\i=\endidx\else\hspace{\imggap}\fi%
            }%
        }\par\nointerlineskip%
        % --- Vertical gap between pairs (except after the final pair) ---
        \ifnum\r<8 \vspace{6pt}\fi%
    }%
    \endgroup
    \vspace{6pt}
    \caption{We present the generated images by Stable Diffusion 1.5 (first rows) and Stable Diffusion 2.1 (second rows) for the decade from 1991-2000. In contrast to the previous figure, here Stable Diffusion 1.5 seems to lead to much less hallucinated content for the years with more abstract artwork.}
    \label{fig:generated_samples_v1_v290}
\end{figure}

\subsection{Augmented Reality}
\label{sec:ar}
The augmented reality component was developed as a standalone and static web application based on html5 and javascript. We used the MindAR library\footnote{\url{https://github.com/hiukim/mind-ar-js}} to recognize the images and overlay the corresponding videos. Every time the webapp is called it randomly selects one of the videos per painting and then plays that video. In this way, the experience is unique for every user. The code for the webpage was written by Gemini 2.5 and the trickiest part was to convince Gemini to force modern mobile phone browsers to automatically play the videos.

\begin{figure}[htbp]
    \centering
    \setlength{\lineskip}{1pt}
    \setlength{\parskip}{0pt}
    \begingroup
    % Calculate exact horizontal gap so 5 images span the line
    \setlength{\imggap}{\dimexpr(\textwidth - 5\dimexpr0.180\textwidth\relax)/4\relax}%

    % Row 1
    \noindent
    \makebox[\textwidth][s]{%
        \includegraphics[width=0.180\textwidth]{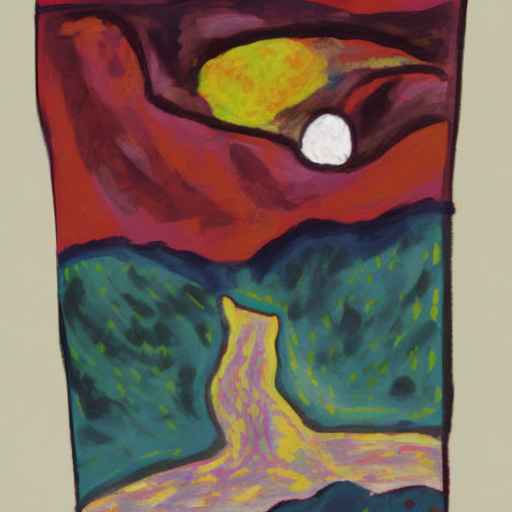}\hspace{\imggap}%
        \includegraphics[width=0.180\textwidth]{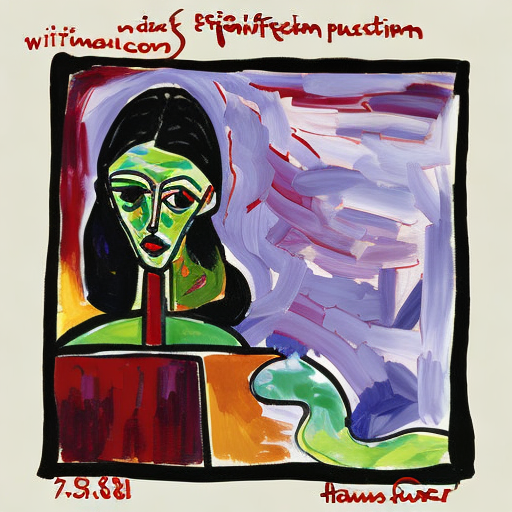}\hspace{\imggap}%
        \includegraphics[width=0.180\textwidth]{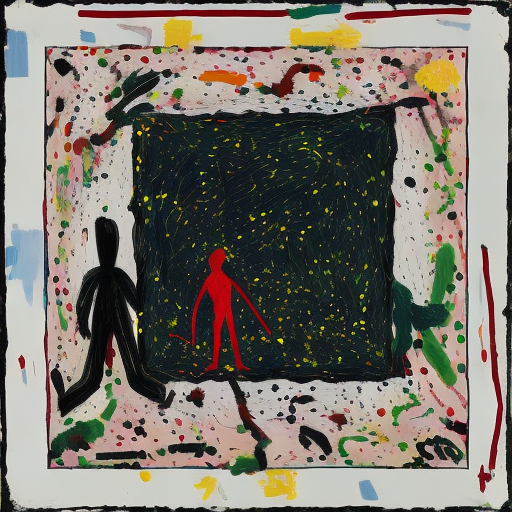}\hspace{\imggap}%
        \includegraphics[width=0.180\textwidth]{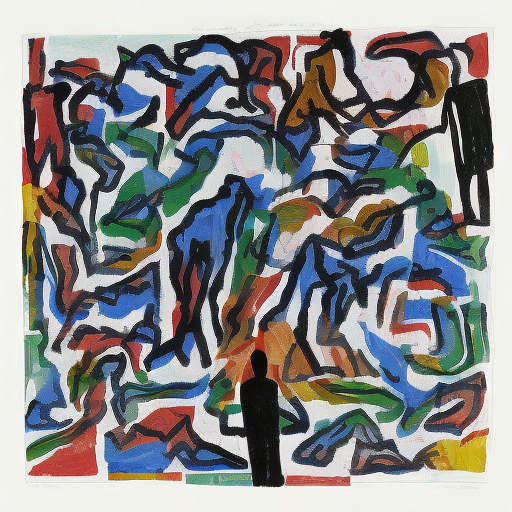}\hspace{\imggap}%
        \includegraphics[width=0.180\textwidth]{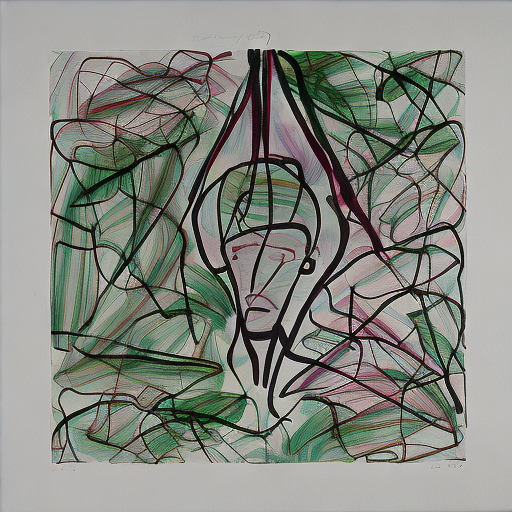}%
    }\par\nointerlineskip
    \vspace{6pt}%

    % Row 2
    \noindent
    \makebox[\textwidth][s]{%
        \includegraphics[width=0.180\textwidth]{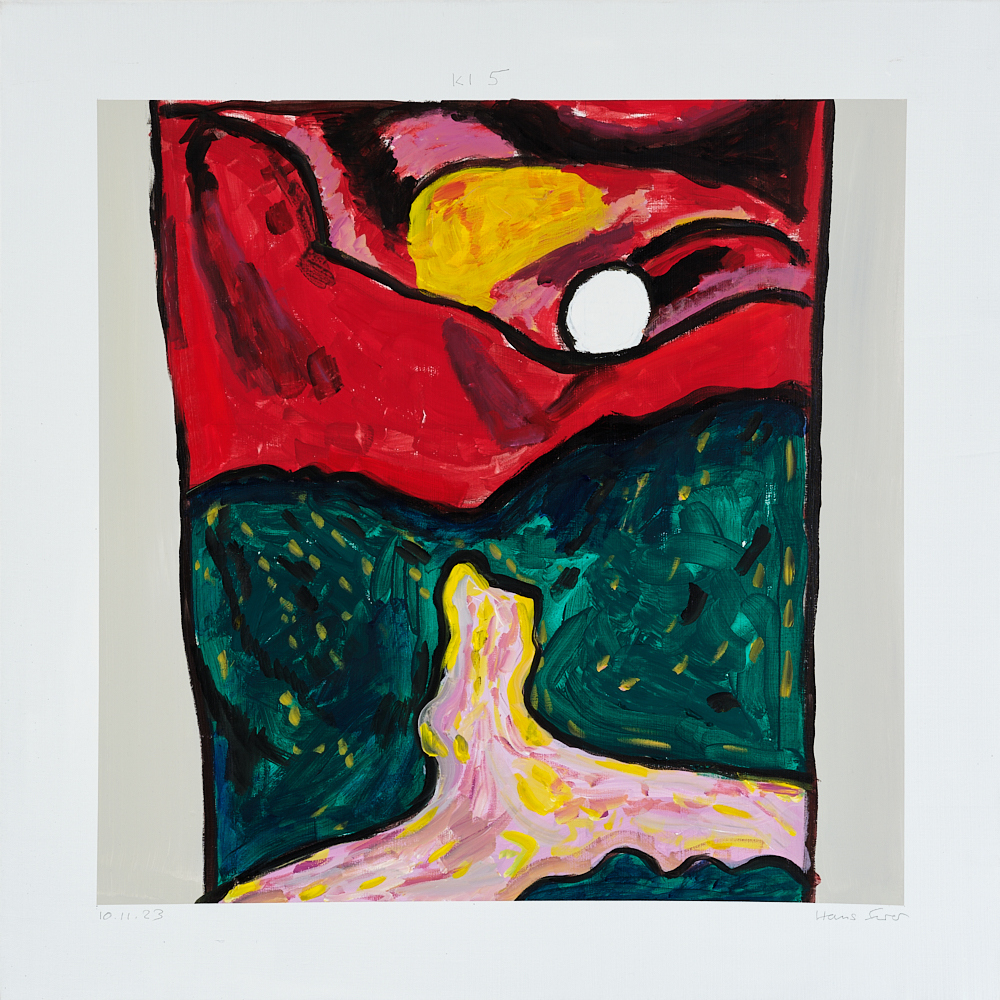}\hspace{\imggap}%
        \includegraphics[width=0.180\textwidth]{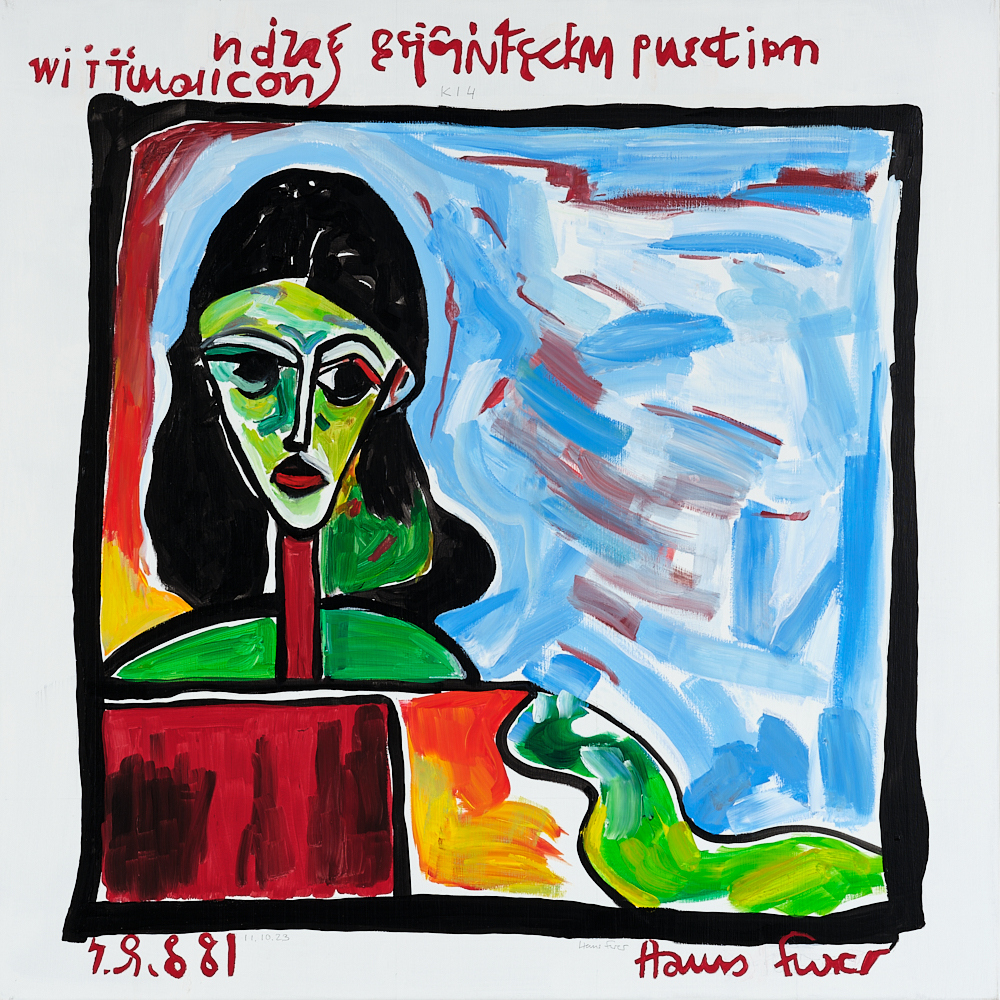}\hspace{\imggap}%
        \includegraphics[width=0.180\textwidth]{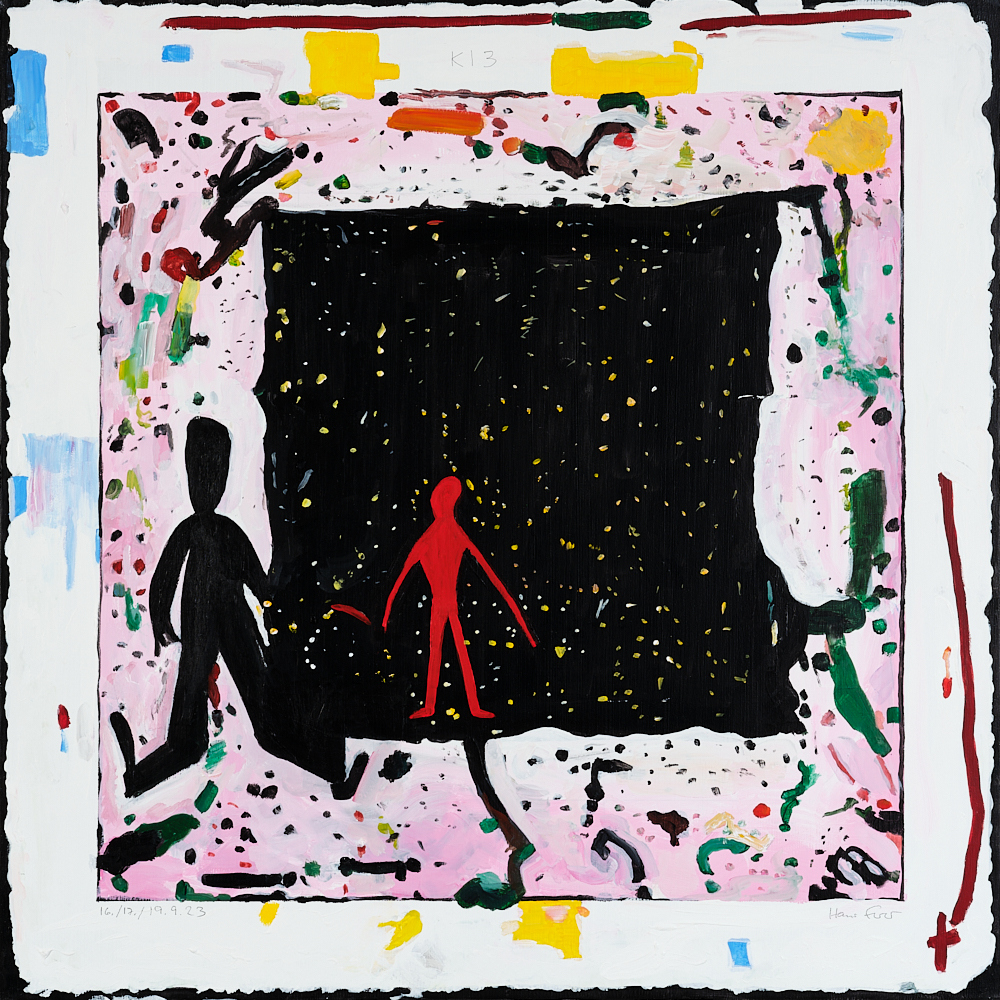}\hspace{\imggap}%
        \includegraphics[width=0.180\textwidth]{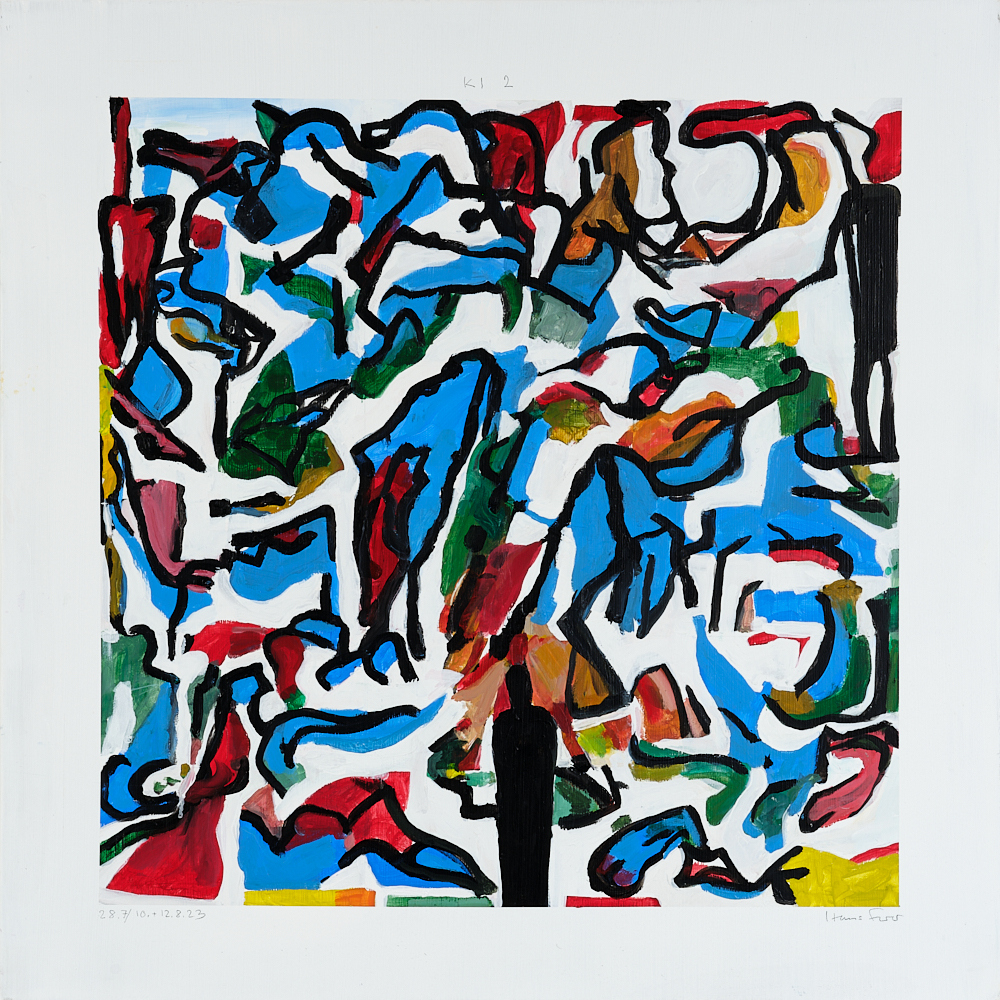}\hspace{\imggap}%
        \includegraphics[width=0.180\textwidth]{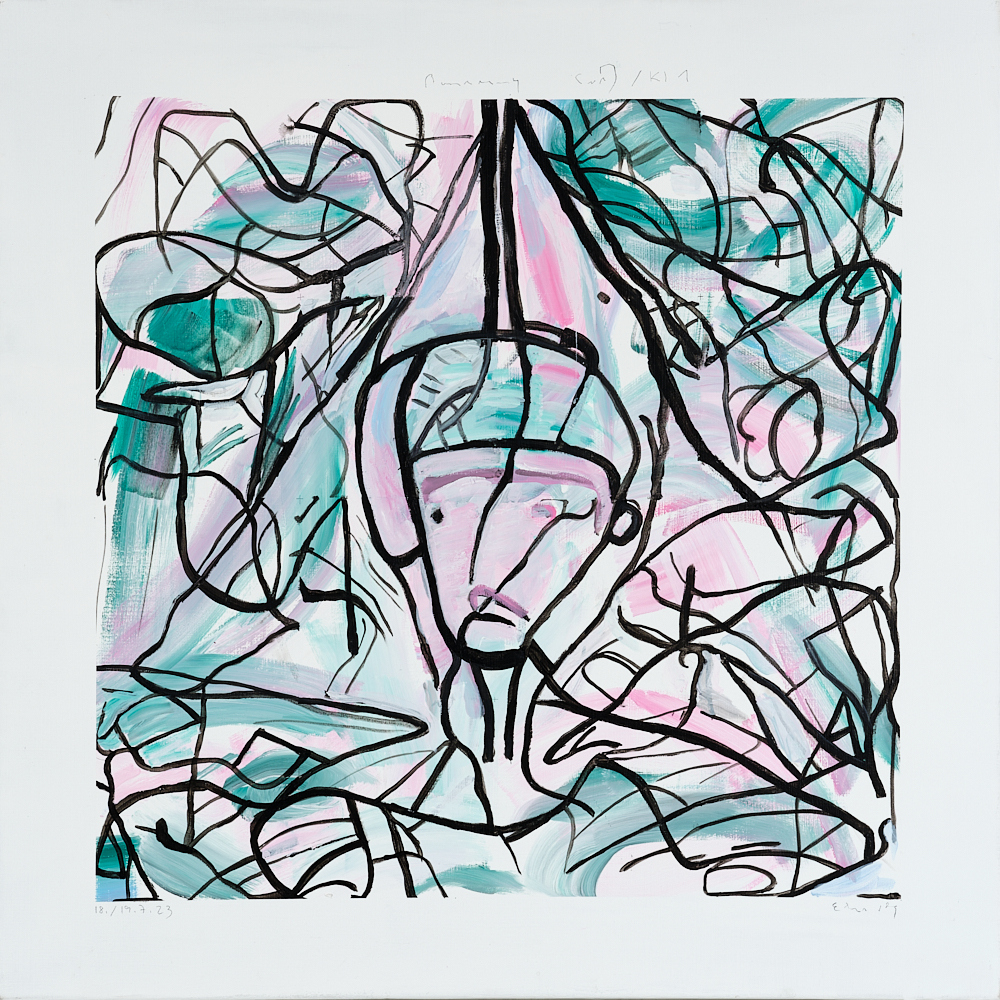}%
    }\par\nointerlineskip
    \vspace{6pt}%

    % Row 3
    \noindent
    \makebox[\textwidth][s]{%
        \includegraphics[width=0.180\textwidth]{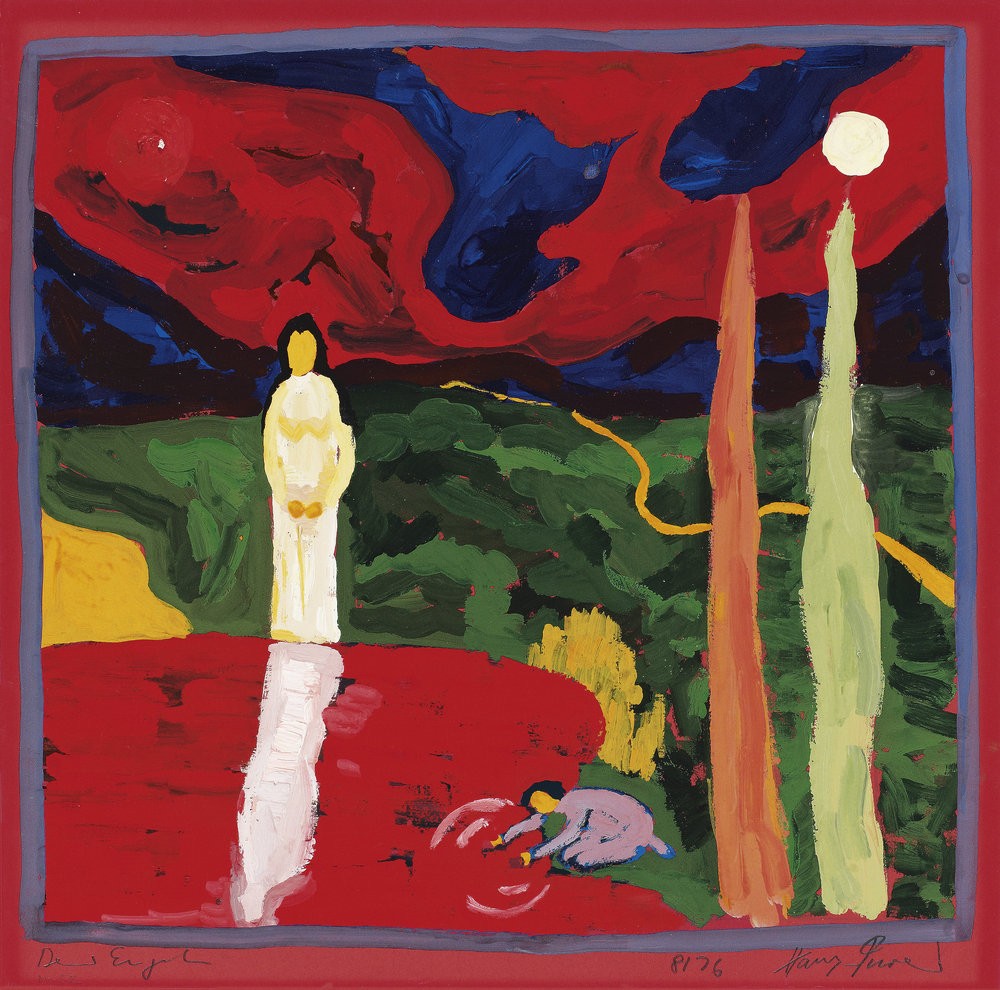}\hspace{\imggap}%
        \includegraphics[width=0.180\textwidth]{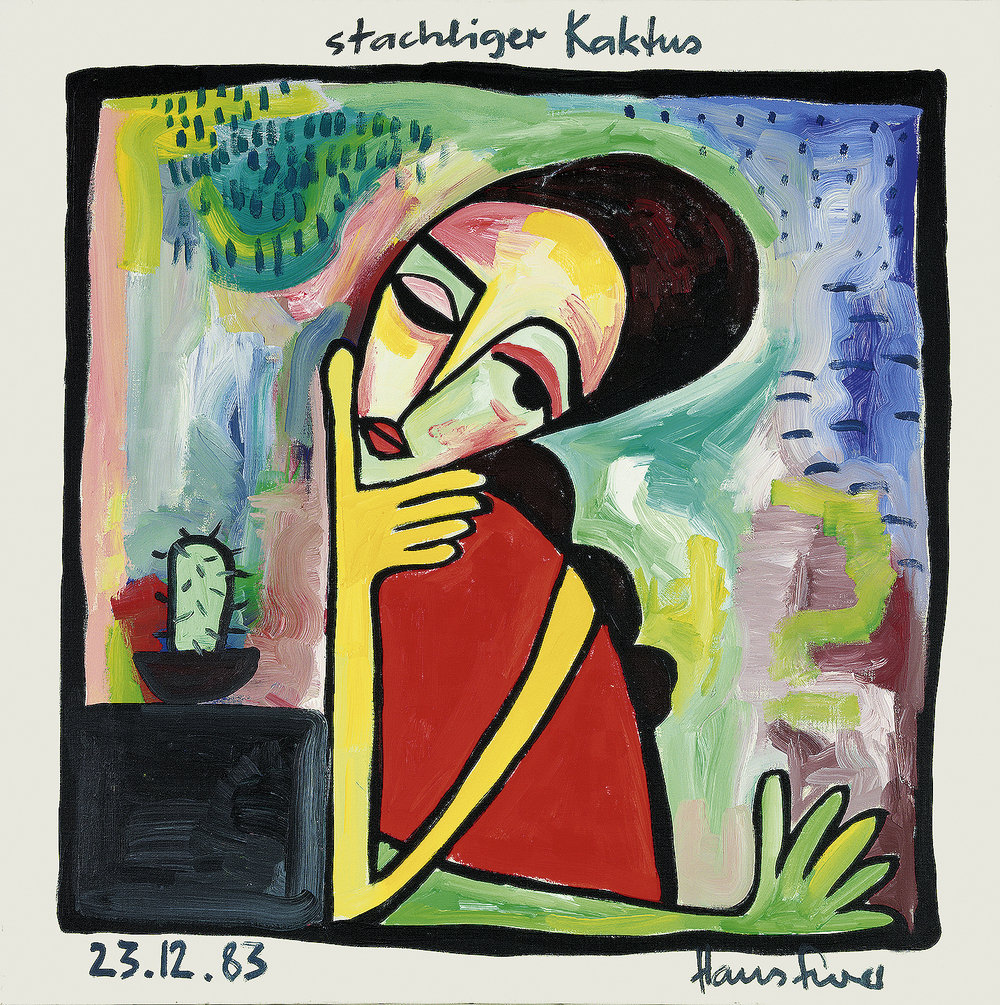}\hspace{\imggap}%
        \includegraphics[width=0.180\textwidth]{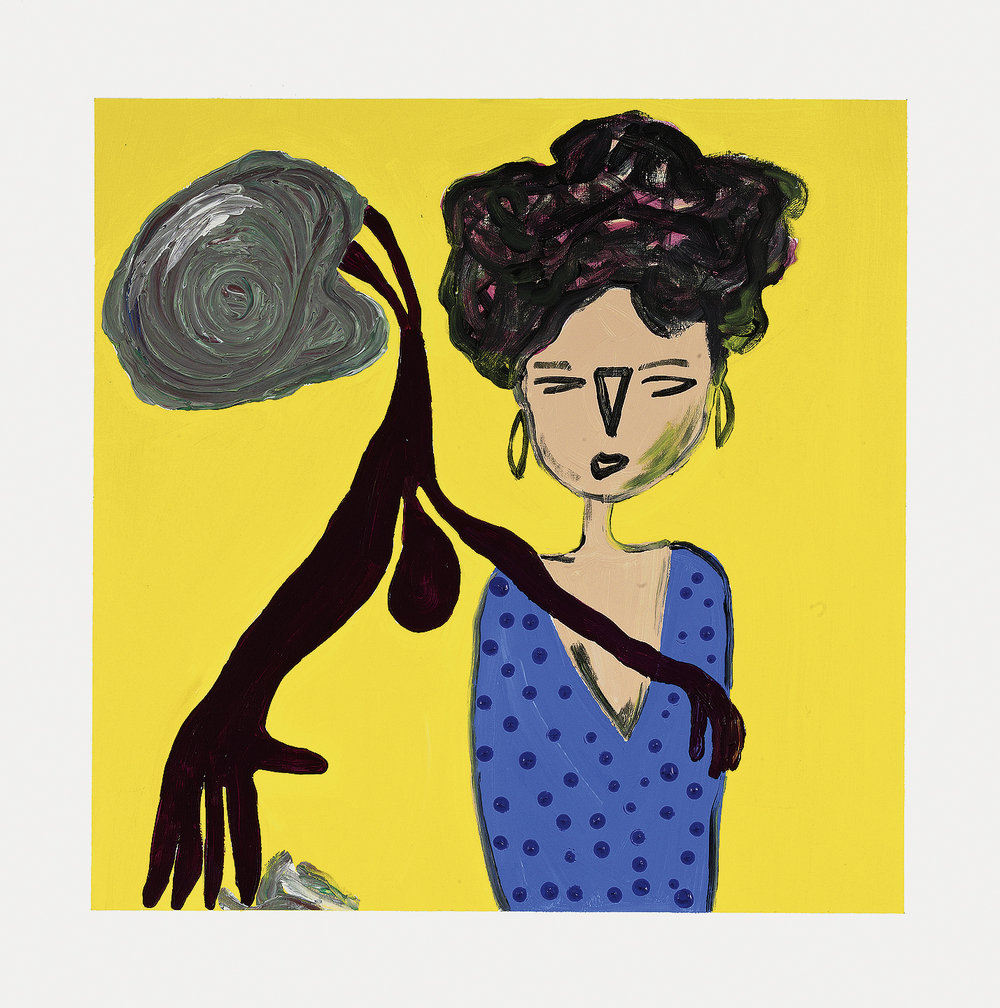}\hspace{\imggap}%
        \includegraphics[width=0.180\textwidth]{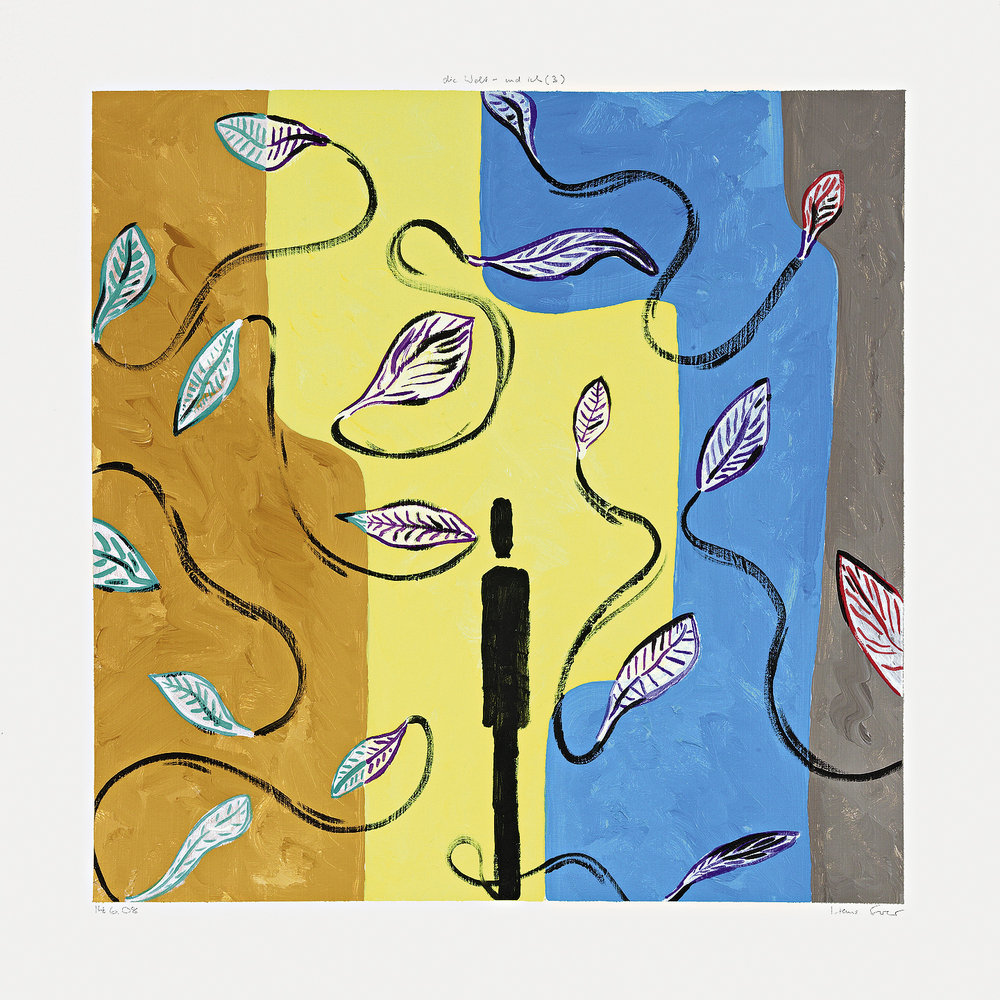}\hspace{\imggap}%
        \includegraphics[width=0.180\textwidth]{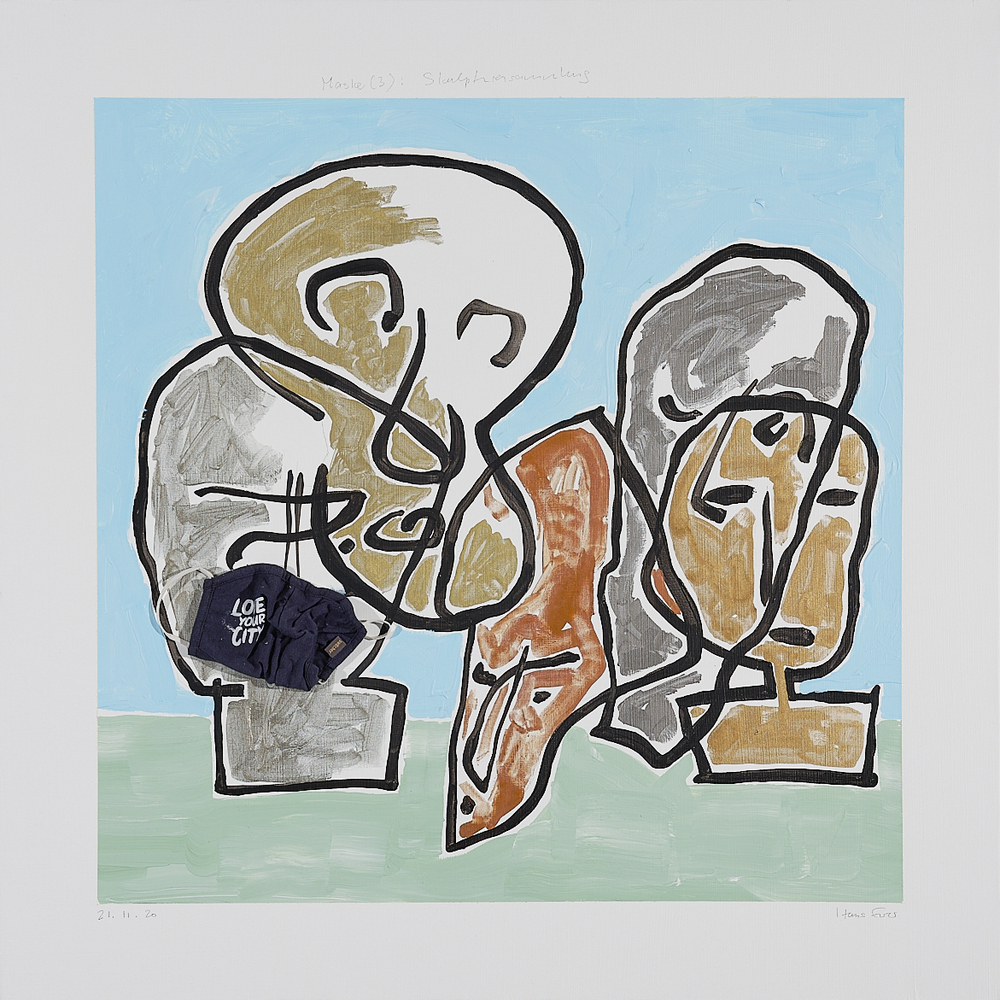}%
    }\par\nointerlineskip
    \vspace{6pt}%

    % Row 4
    \noindent
    \makebox[\textwidth][s]{%
        \includegraphics[width=0.180\textwidth]{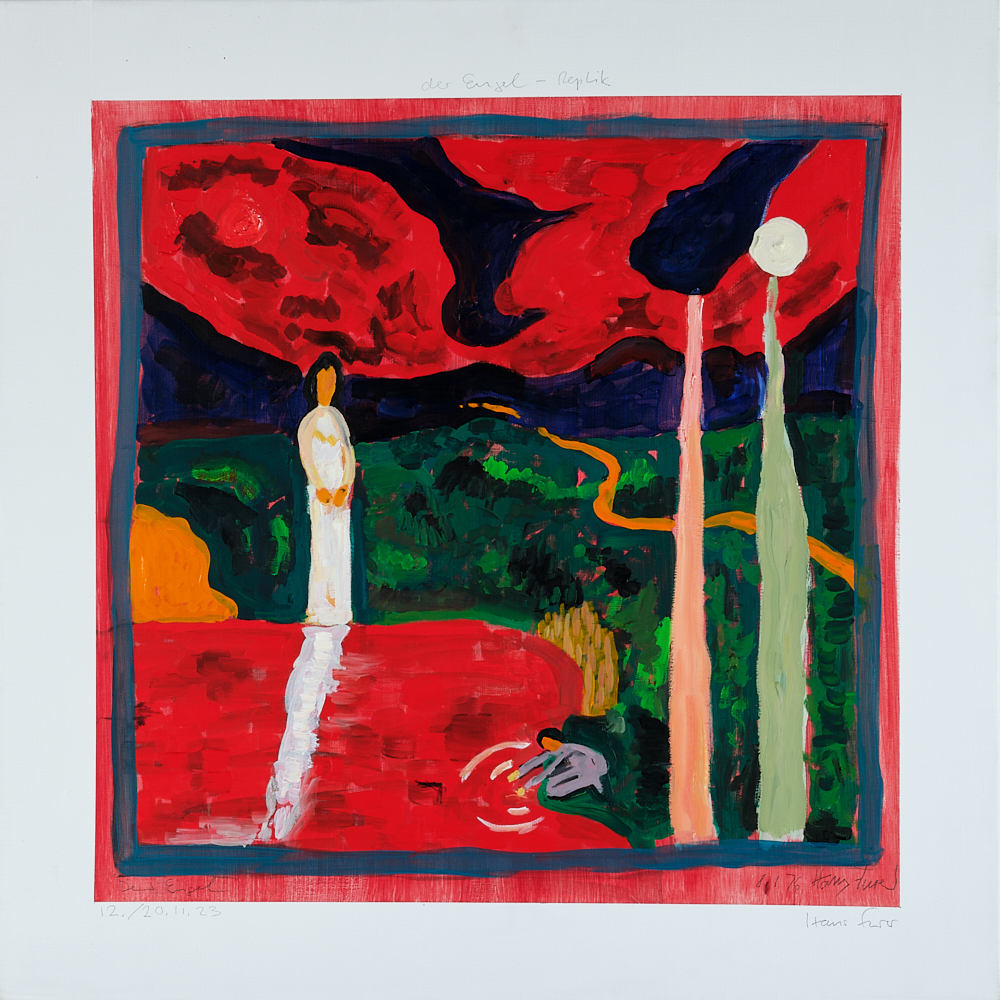}\hspace{\imggap}%
        \includegraphics[width=0.180\textwidth]{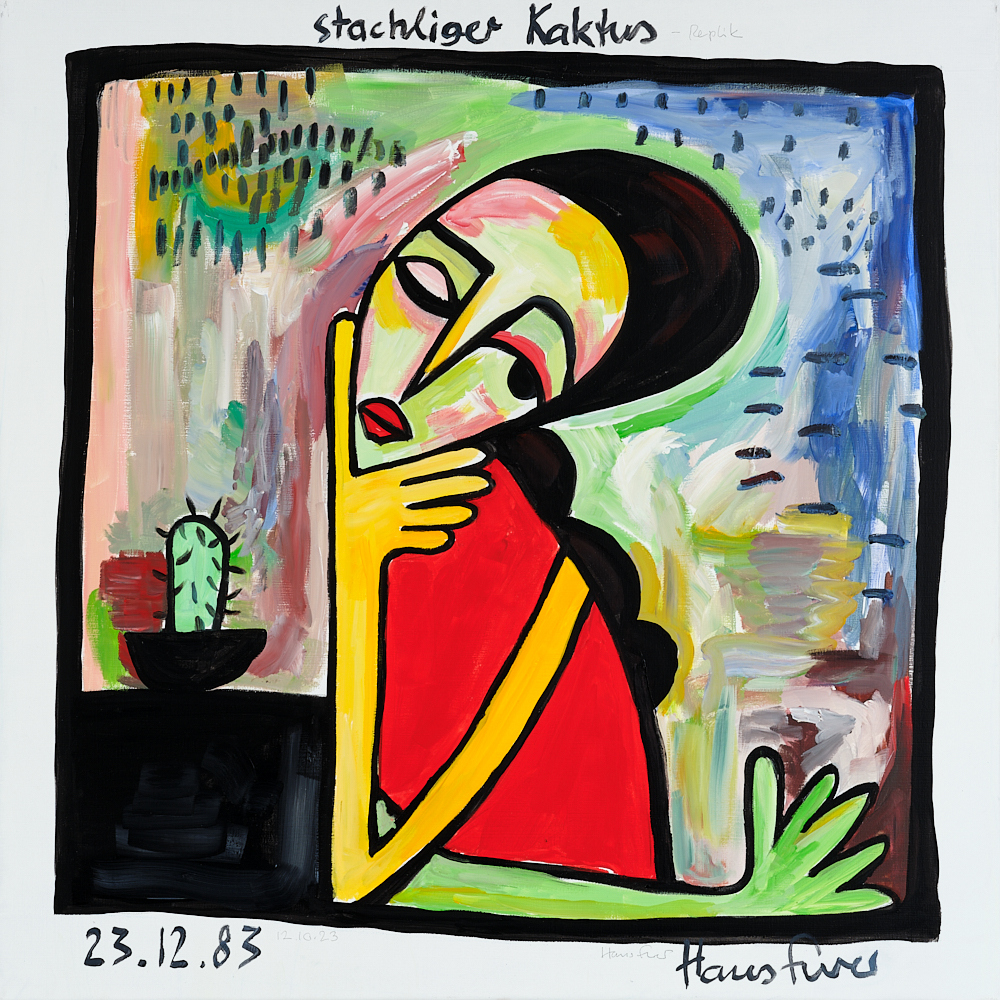}\hspace{\imggap}%
        \includegraphics[width=0.180\textwidth]{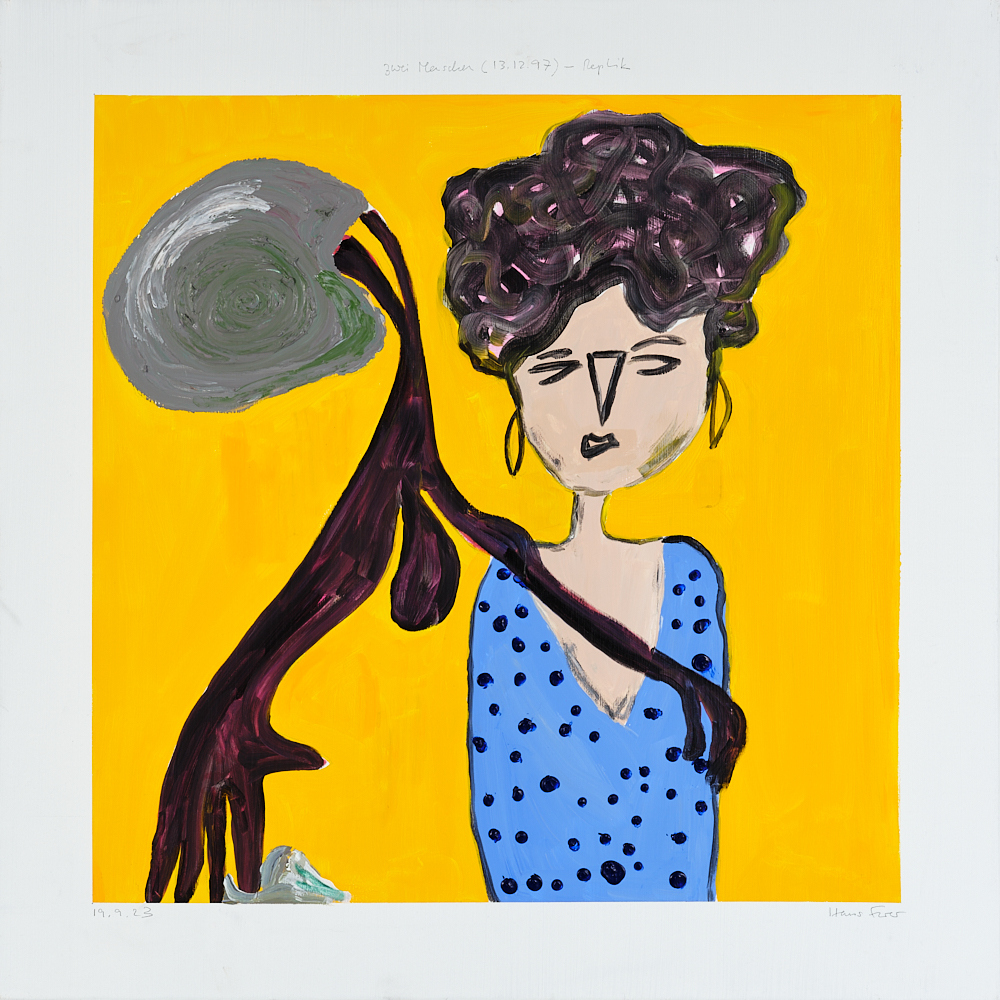}\hspace{\imggap}%
        \includegraphics[width=0.180\textwidth]{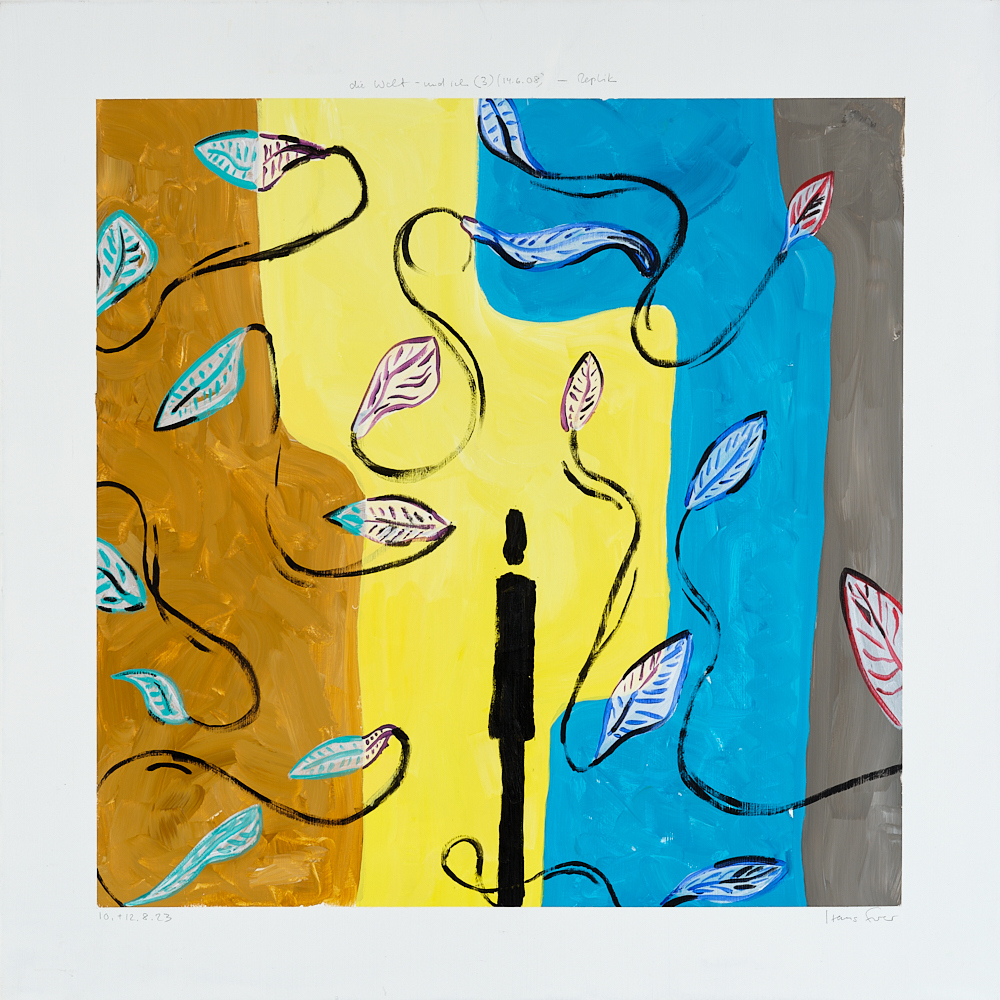}\hspace{\imggap}%
        \includegraphics[width=0.180\textwidth]{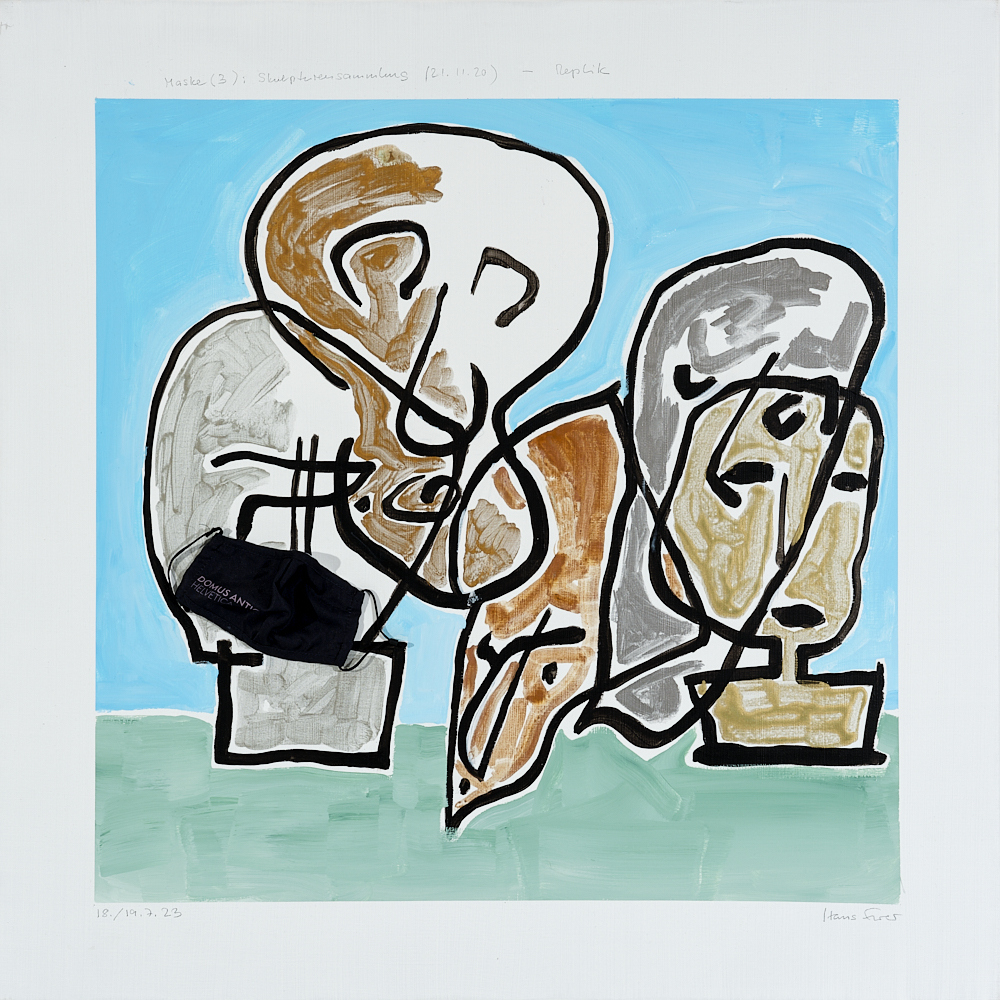}%
    }\par
    \endgroup
  \caption{The 5 images generated through our finetuned diffusion model are shown in the first row. The second row shows photographs of the resulting paintings drawn by the artist. The third row shows for each generated piece the painting from the training data that was selected to be paired with the generated piece. The last row shows the replica of this original work such that both the piece generated with AI and the artist made piece are both executed with the same manual skills. The paintings are ordered chronologically by decade.}
  \label{fig:gallery}
\end{figure}

\section{Experiments and Results}
\label{sec:experiments}

We performed purely observational experiments rather than extensive qualitative and quantitative evaluations with metrics regarding style, image or video quality. This section therefore can rather be seen as a report of how the artwork came along, rather than a comparison to state-of-the-art techniques. We would also like to highlight, that the used methods for generation, fine-tuning and video generation have, in the 3 years since the project was performed, all found successors that would likely lead to different results which might be superior depending on the metric.

\label{sec:imagegen}
\begin{figure}[htbp]
    \centering
    \setlength{\lineskip}{1pt}
    \setlength{\parskip}{0pt}
\begingroup
    % Calculate exact horizontal gap so 7 images span the line
    \setlength{\imggap}{\dimexpr(\textwidth - 7\dimexpr0.130\textwidth\relax)/6\relax}%

    \noindent
    \makebox[\textwidth][s]{%
        \includegraphics[width=0.130\textwidth]{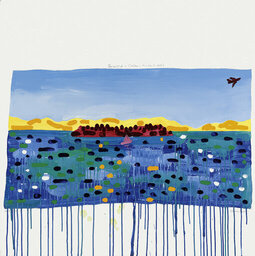}\hspace{\imggap}%
        \includegraphics[width=0.130\textwidth]{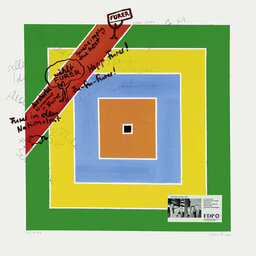}\hspace{\imggap}%
        \includegraphics[width=0.130\textwidth]{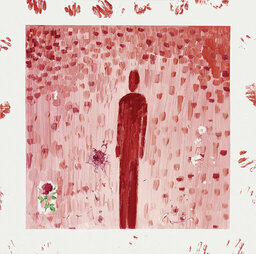}\hspace{\imggap}%
        \includegraphics[width=0.130\textwidth]{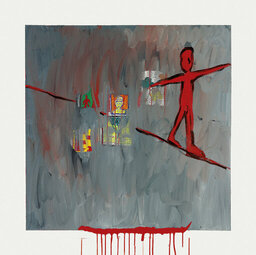}\hspace{\imggap}%
        \includegraphics[width=0.130\textwidth]{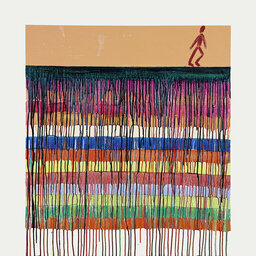}\hspace{\imggap}%
        \includegraphics[width=0.130\textwidth]{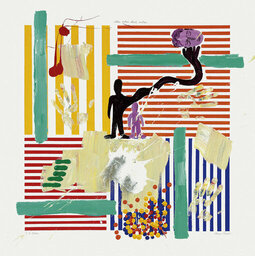}\hspace{\imggap}%
        \includegraphics[width=0.130\textwidth]{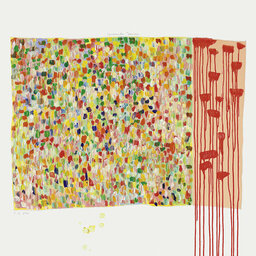}}%
    \endgroup
    \caption{The artist mentioned that the third generated piece shows, that the generative model does not \textit{understand} his work as the model did not leave a 10 cm white border as he usually does for his artwork. A close look at the training data however shows that this is not something the generative model came up with, but is supported by the training data.}
    \label{fig:border}
\end{figure}
\subsection{Image Generation}
We finetuned model variants on 5 different training sets, always on one decade of artwork of Hans Furer. We experimented with both Stable Diffusion 1.5 as well as Stable Diffusion 2.1. resulting in a total of 10 model variants. The resulting pieces for the decade from 1981-1990 and 1991-2000 can be seen in Figure~\ref{fig:generated_samples_v1_v280} and Figure~\ref{fig:generated_samples_v1_v290} respectively. The style of Hans Furer changes quite a bit over time and we observed that the two different models behave differently and depending on the decade generated artwork of one of the models was favored. Since both models share the same basis images generated with the same seed often share quite some similarity, which enables a direct comparison between the models. For the decade from 1981-1990, where Hans Furer drew many faces and bodies the newer model leads to much more detail and makes the images less blurry. For the decade from 1991-2000, which contains much more abstract images in the training data, the newer model seems to struggle more. The key observable difference from our perspective is that the newer model tends to have more texture-like patterns, which could either be a bias inherited by the difference in training data, or a case of hallucination by the model. In contrast, the artwork of Hans Furer is often more calm in the background. For each decade, one images was manually chosen by Bernhard Egger and Hans Furer based on personal preferences. This piece was then transferred to canvas and paired with a piece from the training data that was selected by the artist. The final selection of generated artworks is shown in Figure~\ref{fig:gallery}. This Figure not only shows the final result but also gives a gist on the accuracy at which the artist transcribed the images, including e.g. mistakes the generative model made regarding the text, signature or dates. We however also perceive that the painted artwork appears more vivid.
The third piece, where the generative model did draw into the border which is usually left blank by the artist, again made us think about hallucinations by the model, however we realized that this is actually supported by the training data as can be seen in Figure~\ref{fig:border}.

\subsection{Video Generation}
\label{sec:videogen}
\begin{wrapfigure}[18]{r}{0.4\textwidth}
    \vspace{-0.88cm}
\includegraphics[width=\linewidth, keepaspectratio]{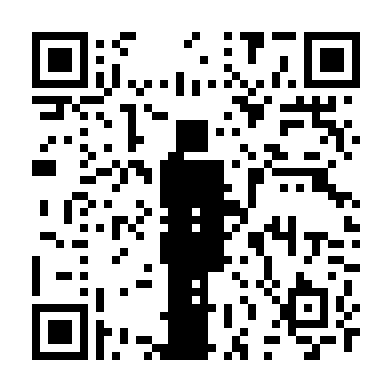}
  \caption{Use this QR code, leading to \href{https://eggerbernhard.ch/AIARt}{eggerbernhard.ch/AIARt} to animate the paintings in Figure~\ref{fig:gallery} (generated by QArt Coder \url{http://research.swtch.com/qr/draw}.}
  \label{fig:qr}
\end{wrapfigure}

Already in 2023 when the exhibition started we regularly tested generative video models for their creative power, but they did not produce favorable animations. In most cases the video model generated a video that was performing simple translation or zooming or generated random other objects in front of the canvas (e.g. humans walking to occlude the image). Our conclusion at that time was that the distribution of the training data is too narrow and walkthroughs of museums or rooms with paintings at walls might lead exactly to that bias. Even with prompt engineering we did not succeed in steering the generation in a favorable direction. Our conclusion at that time was, that the models did not yet generalize well enough to animate abstract image content. Whenever new models (mostly commercial) were released, we gave them a try and in summer 2025 suddenly we started to see the abstract image components to start moving. The first models that we successfully tested regarding animating the artwork in the AI-Experiment were Kling~2.1, Veo~3, Wan~2.2, Hailuo~2. We quickly realized that a large component of the video generation are the actual prompts. Whilst the prompts are by all the used models rewritten by a language model before generation of a video the initial prompt still heavily steers the animation. In contrast to the image generation we did not finetune the models but guide them with an initial frame (sometimes also with another image as end frame) as well as a prompt. The generation of the finally selected videos was crowd-sourced in form of a mini-hackathon, a small event in our lab which attracted eight master and doctoral students that worked on engineering prompts to generate favorable animations with all state-of-the-art video models at that time. The result can be experienced by scanning the QR code in Figure~\ref{fig:qr} and watch the animations behind Figure~\ref{fig:gallery}.

\begin{figure}[htbp]
    \centering
    \setlength{\lineskip}{1pt}
    \setlength{\parskip}{0pt}
    \begingroup
    % Calculate exact horizontal gap so 3 images span the line
    \setlength{\imggap}{\dimexpr(\textwidth - 3\dimexpr0.310\textwidth\relax)/2\relax}%

    \noindent
    \makebox[\textwidth][s]{%
        \includegraphics[width=0.310\textwidth]{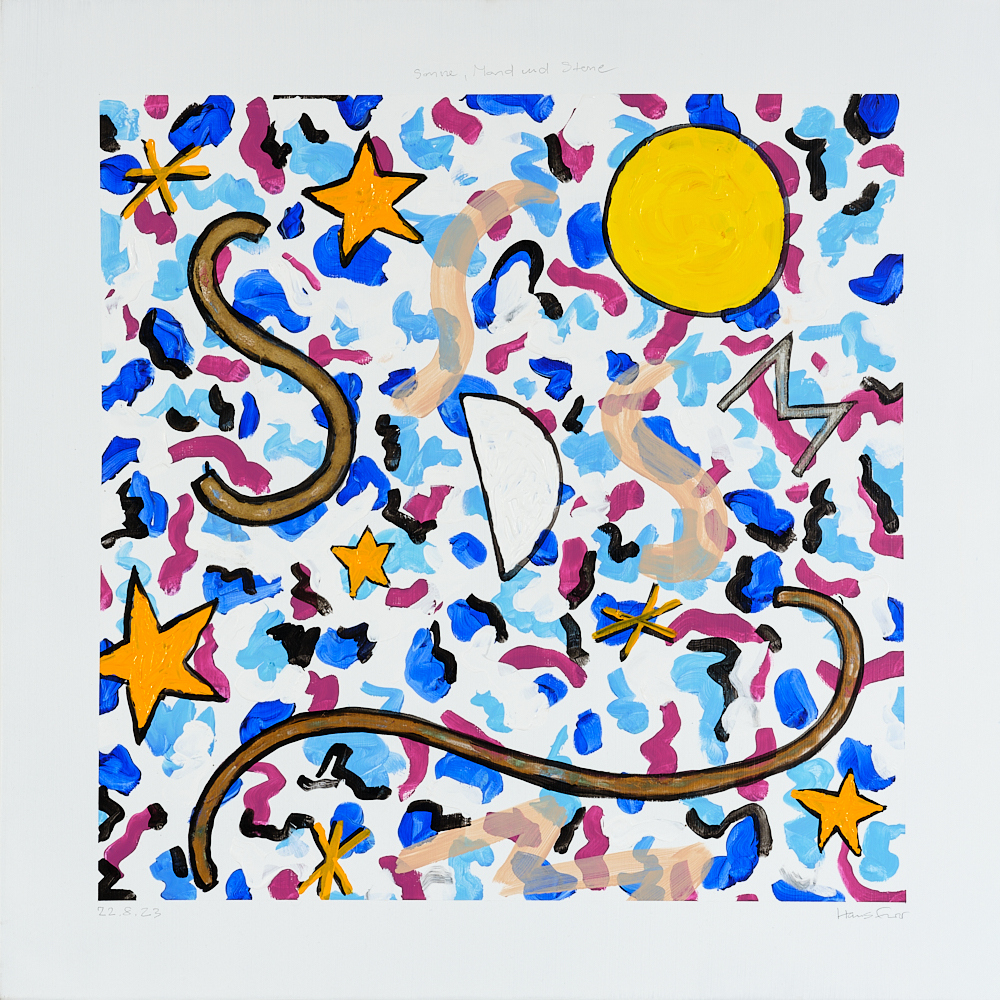}\hspace{\imggap}%
        \includegraphics[width=0.310\textwidth]{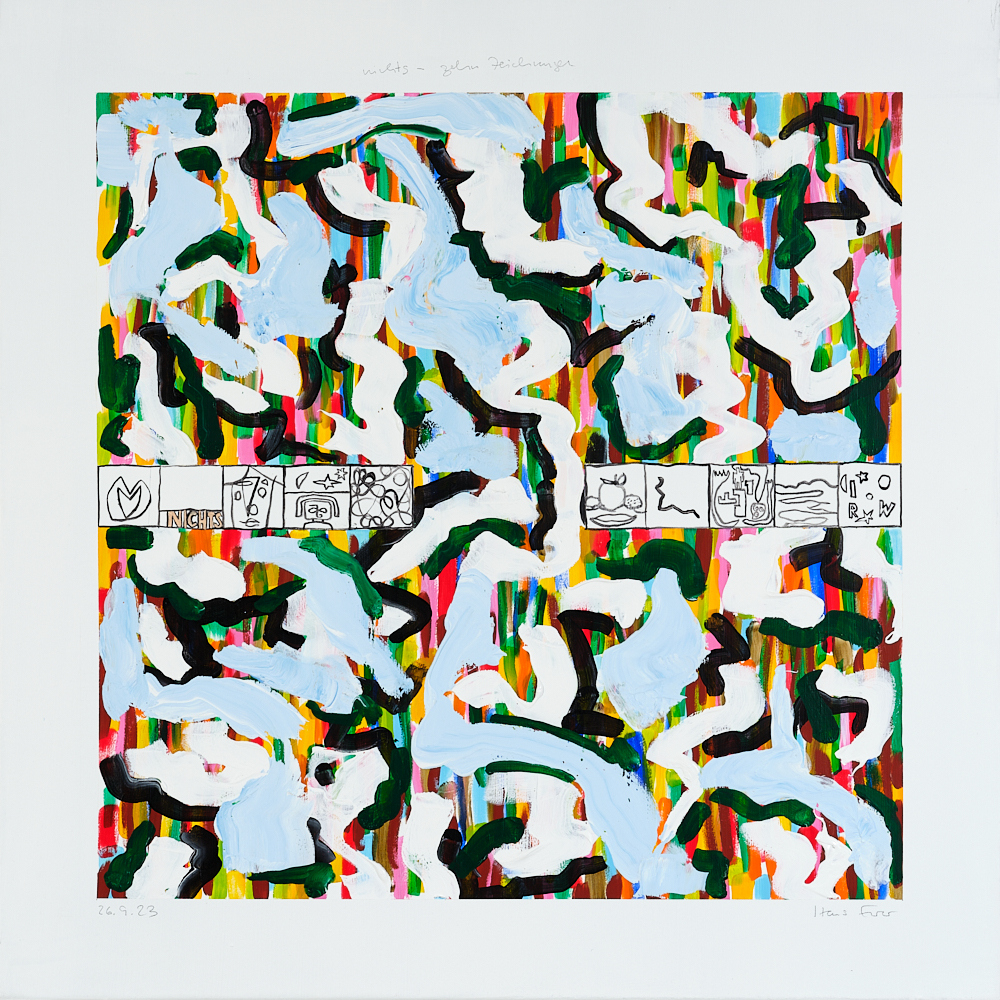}\hspace{\imggap}%
        \includegraphics[width=0.310\textwidth]{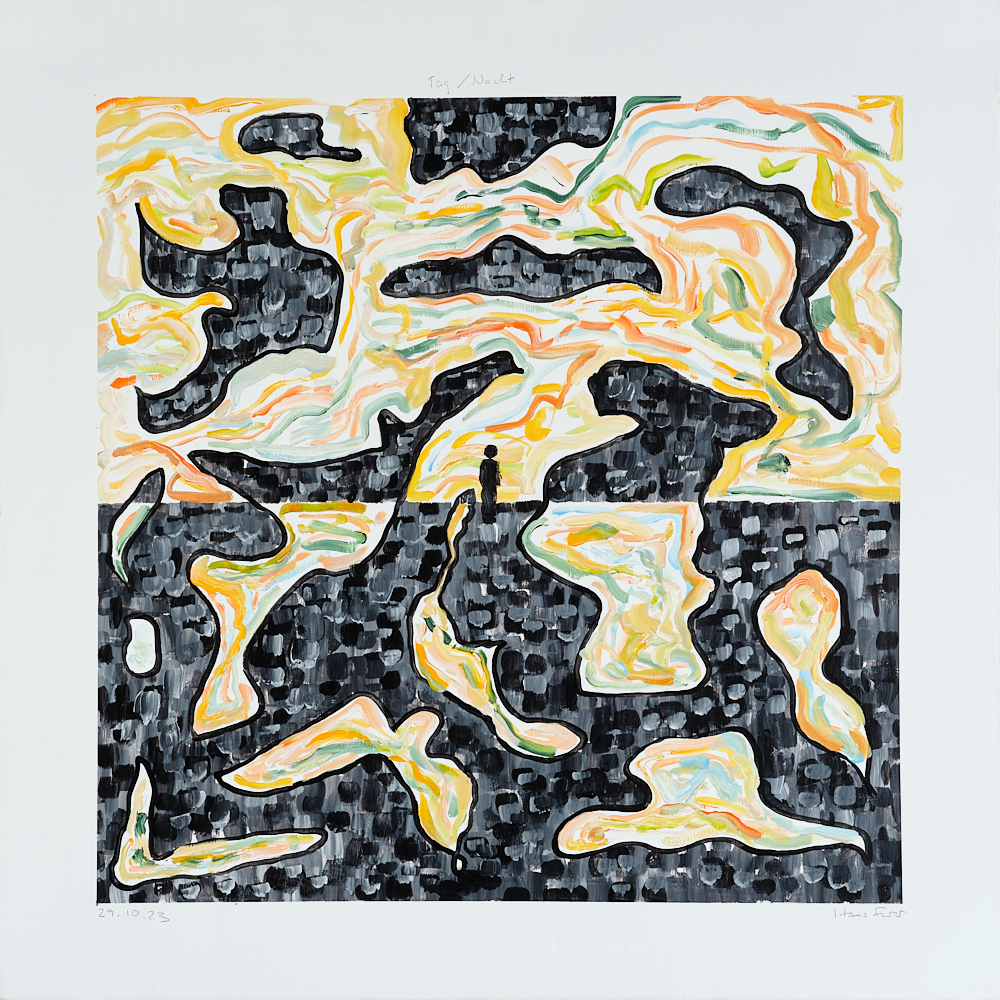}%
    }\par
    \endgroup
    \caption{Following the first two pieces of the AI-Experiment the artist was fascinated by the patterns the generative model made up in the background. Whilst a computer scientist would describe this as a hallucination, as it is something that is not present in the training data, the artist found it interesting, and it inspired him for his subsequent artwork where he started to experiment with wild backgrounds.}
    \label{fig:inspiration}
\end{figure}

\section{Conclusion}
\label{sec:conclusion}

We presented a set of paintings created in synergy between a generative model, a computer scientist and an artist. The unique component is that the project closes the loop - the model is trained on artwork from one specific artist and the artificially generated result is brought back to canvas by the same artist. This combination is unique and also inspired the artist. Especially the hallucinations of the generative model, namely generating crowded backgrounds instead of uniformly colored backgrounds like Hans Furer regularly did, inspired the artist, who found exactly this element to be creative and in his subsequent artworks he experimented with wild backgrounds as can be seen in Figure~\ref{fig:inspiration}. Whilst the artwork itself triggered many discussions when presented at events of the university, the augmented reality extension adds a component that fascinates many visitors from a different aspect. Besides the underlying technology to be known for a while, it is something many people have not seen yet, and the animation of the artworks comes unexpected. In future work we want to further explore this direction and collaborate with teachers and museums to potentially integrate this into teaching and workshops with a focus on art but also computer science.

\section*{Acknowledgements}
We thank Andreea Ardelean, Timotei Ardelean, Nilofaar Jorkash, Takuma Nishi\-mura, Mert Özer, Julian Stahl, Berke Uysal, Daniel Zieger for generating the videos for the augmented reality extension at a mini-hackathon with their prompts they added a unique touch to each of the results. We thank Tobias  R. Dürring who took the photographs of all the involved artwork. We thank Johannes Brendel as well as the reviewers and program chairs of the AI for Visual Arts workshop for their valuable feedback on the manuscript.

\bibliographystyle{splncs04}
\bibliography{main}
\end{document}